\documentclass[final,3p,times,12pt]{elsarticle}
\usepackage{hyperref}
\usepackage{amsmath}
\usepackage{amsfonts}
\usepackage{graphicx}
\usepackage{algorithm}
\usepackage{algorithmicx}
\usepackage{algpseudocode}
\usepackage{booktabs}
\usepackage{array}
\usepackage{multirow}
\usepackage{xcolor}
\usepackage{colortbl}
\usepackage{titlesec}
\usepackage{epsfig}
\usepackage{epstopdf}
\usepackage{caption}
\usepackage{footmisc}
\usepackage{setspace}
\usepackage{txfonts}
\begin{document}

\begin{frontmatter}

\title{Siamese Attentional Keypoint Network for High Performance Visual Tracking}

\author[hitsz,hit]{Peng Gao}
\author[hitsz]{Ruyue Yuan}
\author[hitsz]{Fei Wang}
\author[hitsz,hit]{Liyi Xiao}
\author[es,jp]{Hamido Fujita}
\author[hitsz]{Yan Zhang}
\address[hitsz]{School of Electronics and Information Engineering, Harbin Institute of Technology, Shenzhen, China}
\address[hit]{School of Astronautics, Harbin Institute of Technology, Harbin, China}
\address[es]{Andalusian Research Institute in Data Science and Computational Intelligence (DaSCI), University of Granada, Granada, Spain}
\address[jp]{Faculty of Software and Information Science, Iwate Prefectural University, Iwate, Japan}

\begin{abstract}
Visual tracking is one of the most fundamental topics in computer vision. Numerous tracking approaches based on discriminative correlation filters or Siamese convolutional networks have attained remarkable performance over the past decade. However, it is still commonly recognized as an open research problem to develop robust and effective trackers which can achieve satisfying performance with high computational and memory storage efficiency in real-world scenarios. In this paper, we investigate the impacts of three main aspects of visual tracking, i.e., the backbone network, the attentional mechanism, and the detection component, and propose a Siamese Attentional Keypoint Network, dubbed SATIN, for efficient tracking and accurate localization. Firstly, a new Siamese lightweight hourglass network is specially designed for visual tracking. It takes advantage of the benefits of the repeated bottom-up and top-down inference to capture more global and local contextual information at multiple scales. Secondly, a novel cross-attentional module is utilized to leverage both channel-wise and spatial intermediate attentional information, which can enhance both discriminative and localization capabilities of feature maps. Thirdly, a keypoints detection approach is invented to trace any target object by detecting the top-left corner point, the centroid point, and the bottom-right corner point of its bounding box. Therefore, our SATIN tracker not only has a strong capability to learn more effective object representations, but also is computational and memory storage efficiency, either during the training or testing stages. To the best of our knowledge, we are the first to propose this approach. Without bells and whistles, experimental results demonstrate that our approach achieves state-of-the-art performance on several recent benchmark datasets, at a speed far exceeding 27 frames per second.
\end{abstract}

\begin{keyword}
Visual tracking \sep Siamese hourglass networks \sep cross-attentional module \sep keypoint detection
\end{keyword}

\end{frontmatter}

\section{Introduction}\label{sec1}

Visual tracking is an important research problem in the field of computer vision. It estimates the position of an arbitrary target object in a video sequence, as long as the target object is specified with a bounding box in the first frame. Especially, high-performance visual tracking approaches with real-time speed are widely pursued in numerous applications, including video surveillance, autonomous driving, pose estimation, and human-computer interfaces. Although considerable progress has been made by many tracking approaches in the last decade, visual tracking is still a challenging problem due to multiple negative scenarios such as occlusions, fast motions, scale variations, and background clutters~\cite{survey2006,survey2011,survey2014,survey2018}.

Most recent tracking approaches are developed based on two successful frameworks. The first one is the discriminative correlation filter (DCF). Thanks to the circular shifting of training samples and fast learning correlation filters in the Fourier frequency domain, DCF demonstrates superior computational efficiency and fairly good tracking accuracy and has attracted increasing attention since MOSSE~\cite{mosse}first exploited it. Later, in order to overcome different kinds of challenges and achieve competitive tracking performance, the advancement in DCF-based tracking is focused on the use of kernel functions~\cite{ecs}, motion information~\cite{flowtrack,csot}, multi-dimensional features~\cite{cnt,kcf}, multi-scale estimation~\cite{dsst,samf}, boundary effects alleviation~\cite{srdcf,bacf}, deep convolutional features~\cite{eco,ccot}, and ensemble combination~\cite{cact,mlcft,mfcmt}. However, most DCF-based trackers employ offline pre-trained convolutional neural networks (CNN) for feature extraction and do not perform stochastic gradient descent (SGD) to online fine-tune the parameters. So that DCF-based trackers benefit very little from end-to-end trainable networks.

In recent years, another popular visual tracking framework occurs, i.e., fully-convolutional Siamese network (SiamNet), which has shown remarkable potential for high-performance visual tracking. These SiamNet-based trackers use a Siamese convolutional network to compare an exemplar image with candidate images of the same size, and hence bypass the online fine-tuning problem~\cite{sint,siamfc,cfnet,cfcf,lcsnt}. Several recent extensions~\cite{siamrpn,dasiamrpn} exploit a SiamNet for feature extraction, and a Region Proposal Networks (RPN)~\cite{fastrcnn,fasterrcnn} for classification and detection, have yielded excellent results on various benchmark datasets.

Unfortunately, the visual tracking community is still plagued by several problems. First, existing backbone networks~\cite{alexnet,vgg,resnet} applied in visual tracking are always pre-trained for other different tasks such as ImageNet object classification and detection~\cite{imagenet}, which are not adequate for effective tracking. In these networks, feature maps extracted from high-level convolutional layers always have low resolutions, which are not sufficient for accurate localization. In contrast, high-resolution feature maps extracted from lower layers are ineffective to distinguish targets with different attributes or categories. Moreover, these pre-trained networks are not lightweight enough for real-world applications. Existing lightweight networks, such as MobileNet~\cite{mobilenet}, SqueezeNet~\cite{squeezenet}, and Xception~\cite{xception}, have demonstrated fast and effective performance with slightly accuracy degradation on object classification and detection tasks compared to popular networks. Still, they cannot fully take advantage of the benefits of deeper and wider architectures to enhance capabilities for target object representation in visual tracking~\cite{cir}. Therefore, how to effectively make trade-off between computational complexityand strong representation capability of the backbone network is critical for visual tracking. Second, it is well known that intermediate relationships of different convolutional layers demonstrate the true potential for improving the representation and generalization capability~\cite{senet,cbam}, but only a few approaches~\cite{flowtrack,rasnet,lsart,rar} efficiently take full use of these relationships in visual tracking. Third, in order to obtain more accurate tracking results, recent trackers~\cite{siamrpn,dasiamrpn} employ RPN for proposing target candidates. However, RPN always requires a large number of anchor boxes~\cite{fastrcnn,fasterrcnn} with redundant hyperparameters, i.e., the number, scales, and aspect-ratios of the candidate proposals, which incur computational and memory storage overload and make these approaches not efficient both in the training and testing stages.

\begin{figure}[t]
\begin{center}
    \includegraphics[width=\linewidth]{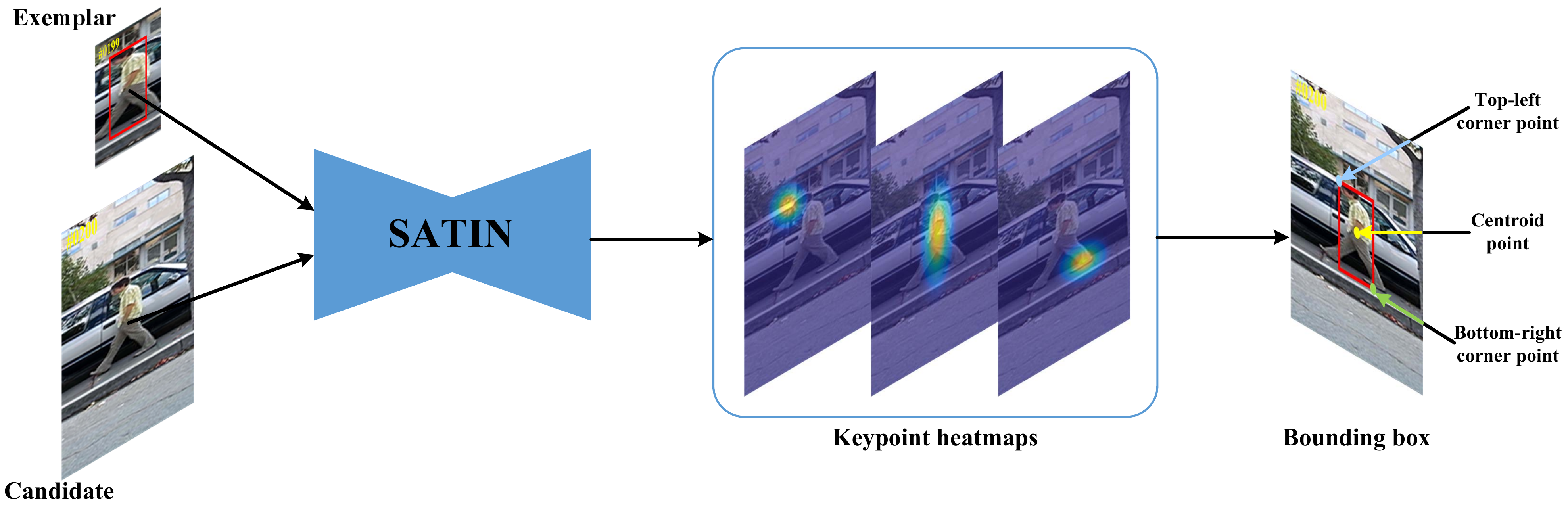}
\end{center}
\caption{Visualization of heatmaps of SATIN. Tracking result on the $200^{th}$ frame of the sequence \emph{David3} is presented, with the top-left corner point, the centroid point and the bottom-right corner point marked.}
\label{fig:heatmaps}
\end{figure}

To address the above issues, we develop the \emph{Siamese Attentional Keypoint Network}, dubbed SATIN, for high-performance visual tracking in an end-to-end fashion. It consists of three components, i.e., a Siamese lightweight hourglass network for feature extraction, a cross-attentional module for adaptive feature refinement, and a keypoints detection module for object detection and localization. As the backbone network of SATIN, the Siamese lightweight hourglass network is symmetric, consisting of (a) a series of convolutional layers for feature map resolution reductions, (b) a series of upsampling layers to raise the low-resolution feature maps to the original resolution and (c) a set of skip layers to bring back original information. We also shrink the network by reducing the number of hyperparameters to make it simple and fast~\cite{mobilenet}. Therefore, our backbone network takes advantage of the benefits of the repeated bottom-up and top-down framework and captures more contextual information at multiple scales on the input images. Moreover, to boost the representation power of SATIN, we investigate the impact of spatial and channel-wise attention mechanisms. It is well known that visual attention plays the most critical role in many computer vision tasks. We align our motivation to the new convolutional block attention module (CBAM)~\cite{cbam} and develop a novel cross-attentional module consisting of a multi-layer perceptron and a single-layer perceptron to highlight informative representations and suppress the background noises. Since low-level geometric information is effective for target object localization, and high-level semantic information is sufficient for target object discrimination, the spatial and channel-wise attention maps can be computed separately on different level information. Further, motivated by CornerNet~\cite{cornernet}, we propose a unified detection architecture, which traces the target object using a set of keypoints, as shown in Fig.~\ref{fig:heatmaps}. It detects the top-left corner point, the centroid point, and the bottom-right corner point of the target bounding box using three separate detection module. The centroid point detection module is similar to SiamFC~\cite{siamfc}. Each corner point detection module has a localization branch and an offset prediction branch, starting with its corresponding pooling layer, namely, \emph{corner pooling}~\cite{cornernet}, to overcome the limitation of the receptive field, and enhance the capability to localize corner points of the bounding box. These corner points are determined by searching for the highest scores on the heatmaps, and their locations are adjusted by the offsets. Compared with recent RPN methods, our keypoints detection module is simple, which eliminates the need for designing anchor boxes, and therefore significantly reduces computational and memory storage overhead.

Our main contributions can be summarized as three folds. (a) We design a Siamese lightweight hourglass network as the backbone, which can capture more contextual information at multiple scales on the input images. (b) We develop a novel cross-attentional module to selectively highlight meaningful information, and to boost the representation power of feature maps. (c) We trace the target object by detecting the top-left corner, the centroid, and the bottom-right corner of the target bounding box to avoid designing anchor boxes or multi-scale pyramids. Finally, our proposed approach achieves consummate performance both in terms of accuracy and robustness on several recent benchmark datasets~\cite{otb2013,otb2015,vot2016,vot2017,vot2018}. To the best of our knowledge, we are the first to formulate visual tracking as a keypoint detection task.

The rest of the paper is organized as follows. Section~\ref{sec:2} briefly reviews related works. Section~\ref{sec:3} illustrates the proposed SATIN tracker. Section~\ref{sec:4} details experiments and discusses results. Finally, Section~\ref{sec:5} concludes this paper with final remarks.

\section{Related works}\label{sec:2}

In this section, we give a brief review of three aspects related to our work: backbone networks, attentional mechanisms, and detection components.

\subsection{Backbone networks}\label{sec:21}

In recent years, CNNs have made significant progress in a wide range of computer vision applications due to their impressive representation abilities. Because of their surprisingly good performance of CNN on object classification and detection, researchers are encouraged to either combine existing CNNs with DCF, or design deep networks in the Siamese framework for high-performance visual tracking. The most popular backbone networks utilized in recent trackers~\cite{ccot,csot,hcf,siamfc,siamrpn,lmsco} are AlexNet~\cite{alexnet}, VGGNet~\cite{vgg}, and ResNet~\cite{resnet}. AlexNet~\cite{alexnet} consists of several convolutional and pooling layers, and it was the first large-scale CNN that had won the ImageNet Large Scale Visual Recognition Challenge (ILSVRC)~\cite{imagenet}. VGGNet~\cite{vgg} stacks numerous small convolutional kernels without using pooling operations, which increases the representation power of the network while reducing the number of parameters. ResNet~\cite{resnet} introduces the skip connection to learn residual information, which makes it more efficient and simple to design deeper architectures.

Moreover, there are some efficient backbone networks introduced by other vision tasks, such as hourglass networks~\cite{hourglass} and FlowNet~\cite{flownet}. However, these networks sometimes are computational and memory expensive for practical computer vision applications. Meanwhile, some lightweight networks~\cite{mobilenet,squeezenet,xception} focus on designing a more efficient architecture to reduce network computation while maintaining excellent performance. Unfortunately, these networks mentioned above are always pre-trained for object classification and detection. Trackers that employ these networks may obtain suboptimal tracking results. The recent trend in visual tracking~\cite{mdnet,cfnet,siamfc} is to design suitable networks for learning object- or category-specific representations and enhance the generalization power to new video sequences. Different from these tracking methods, we aim to design a lightweight CNN as the backbone network, which can learn more contextual features at multiple scales with a simple architecture and small model size.

\subsection{Attentional mechanisms}\label{sec:22}

Attentional mechanisms have been widely utilized in many computer vision tasks, such as image classification~\cite{senet}, object detection~\cite{cbam}, and image semantic segmentation~\cite{panet}. The recent advance of tracking approaches has achieved great success by integrating attentional mechanisms. CSRDCF~\cite{csrdcf} introduces a channel weighted block by constructing a unique spatial reliability map in an online constrained optimization. LSART~\cite{lsart} proposes a cross-patch similarity kernel to encode the similarity scores of all pairs of patches between two samples, and a spatial regularized constraint method to enforce each filter kernel to focus on a specific region of the target. Both RASNet~\cite{rasnet} and FlowTrack~\cite{flowtrack} adopt a squeeze-and-excitation network (SENet)~\cite{senet} to enhance the representation capability of their backbone networks. In these modules, global average-pooling operations are utilized to compute channel-wise attention, and to re-weight the original feature maps based on the pooled features. In contrast to those attentional mechanisms, we design a cross-attentional module in an end-to-end offline learning framework to investigate both spatial and channel-wise relationship between shallow and deep convolutional features, especially their geometric and semantic contributions to global contextual information.

\subsection{Detection components}\label{sec:23}

Most state-of-the-art tracking approaches follow the tracking-by-detection paradigm~\cite{lowrank,dsst,csot,siamrpn}, which treats visual tracking as detection problems. DCF-based trackers~\cite{kcf,dsst} and some SiamNet-based trackers~\cite{cfnet} detect objects by searching the peak value on similarity score maps, which come from the correlation of the candidate feature maps with correlation filters trained on the exemplar feature maps. While other SiamNet-based trackers~\cite{siamfc} straightforwardly cross-correlate exemplar and candidate feature maps. To achieve more accurate tracking in scale variation scenarios, most existing tracking approaches utilize multi-scale pyramids as DSST~\cite{dsst} and SAMF~\cite{samf}. But these improvements in accuracy are gained at the expensive cost in tracking speed. SiamRPN~\cite{siamrpn} and DaSiamRPN~\cite{dasiamrpn} combine RPN~\cite{fasterrcnn} with AlexNet in the Siamese framework to formulate the multiple scale object tracking as proposal classification and bounding box regression problem on correlation feature maps. However, these trackers always require a large number of anchor boxes, and many extra hyperparameters have to be set in advance to generate more suitable proposals. Unlike these detection methods, and motivated by object detection approach in CornerNet~\cite{cornernet}, we trace target object as a set of keypoints, i.e., the top-left corner point, the centroid point and the bottom-right corner point of the target bounding box, therefore mitigate the drawbacks of the above-mentioned detection components.

\section{The proposed approach}\label{sec:3}

\begin{figure}[t]
\begin{center}
    \includegraphics[width=\linewidth]{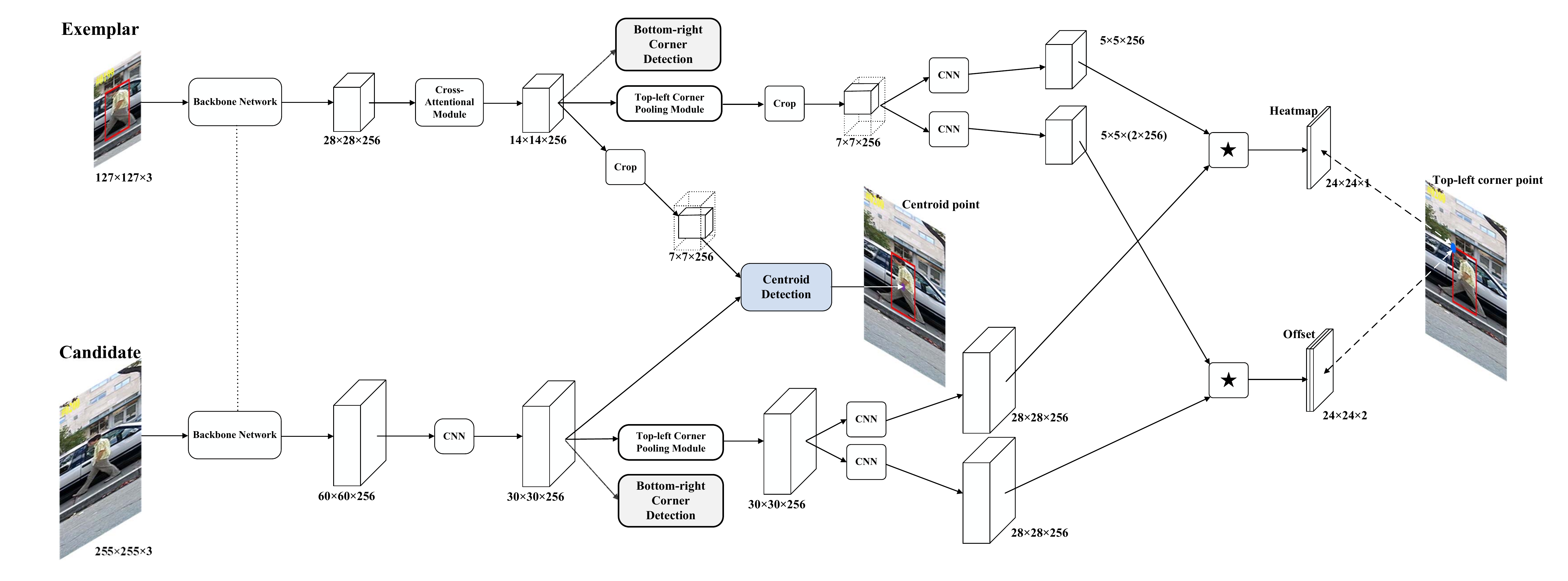}
\end{center}
\caption{The overall framework of our proposed approach. For the sake of clarity, we only show pipelines of the top-left corner (bottom-right corner) point and the centroid point detection.}
\label{fig:arch}
\end{figure}

\subsection{Algorithm overview}\label{sec:31}

In this section, we describe our proposed SATIN in detail. Fig.~\ref{fig:arch} shows the overall framework of SATIN, which consists of three portions: a backbone network for deep feature extraction, a cross-attentional module for representation capability improvement, and a keypoints detection module for keypoints prediction. Given an exemplar image patch $\mathbf{z}$ centered on the target object and a larger candidate image patch $\mathbf{x}$ centered on the previous target location, these two images are fed into a backbone network consisting of two lightweight hourglass networks. Each lightweight hourglass network is designed with a symmetric topology that captures and consolidates both global and local information at multiple scales of the input image through repeated bottom-up and top-down processing. The architecture of our backbone network will be presented in Section~\ref{sec:32}. The backbone network is followed by a cross-attentional module, which is consisted of a multi-layer perceptron and a single-layer perceptron to highlight meaningful information along spatial and channel-wise dimensions, respectively. The spatial attention map is inferred from the low-level geometric features before being fed into the second stage lightweight hourglass network. While the channel-wise attention map is obtained from the high-level semantic feature map, which is the output of the second stage lightweight hourglass network. Both spatial and channel-wise attention maps are then multiplied with the output of the backbone network, therefore taking full advantage of essential attention during tracking. We will illustrate the details of our cross-attentional module in Section~\ref{sec:33}. After that, we exploit the keypoints detection module to trace target objects, which detects the top-left corner point, the centroid point, and the bottom-right corner point of the bounding box simultaneously. Each corner point detection module has its corner pooling strategy and correlates the candidate feature map with the exemplar feature map to obtain a heatmap for corner location prediction and a set of offsets for corner location refinement. The bottom-right corner and the top-left corner are detected similarly. The centroid point is detected as similar to SiamFC~\cite{siamfc}. For the sake of clarity, Fig.~\ref{fig:arch} demonstrates the processes of top-left corner point detection and centroid point detections. We will describe the implementation of the keypoint detection module in Section~\ref{sec:34}.

\subsection{Lightweight hourglass network}\label{sec:32}

It is well acknowledged that powerful target object representations are crucial for accurate and robust visual tracking~\cite{winsty,csot}. Deep features extracted from the off-the-shelf CNNs pre-trained for other visual tasks may be suboptimal for visual tracking, there is still a considerable gap to achieve state-of-the-art tracking performance. Different from CNNs employed in most existing tracking approaches~\cite{siamfc,siamrpn,csot}, we employ a Siamese lightweight hourglass network as the backbone network of SATIN, which is technically designed for visual tracking.

The hourglass network was first introduced into the computer vision community for human pose prediction~\cite{hourglass}. In the original hourglass network, several \emph{hourglass modules} are stacked to preserve low-level geometric representation and high-level semantic information across different resolutions in the repeated bottom-up (from fine to coarse) and top-down (from coarse to fine) processing. Each hourglass module consists of several \emph{residual blocks}, which are the minimal units in the network. The hourglass module first downsamples input features to coarse resolution using a series of convolutional and max-pooling layers to extract semantic information. The pooled features are then upsampled back to the original input resolution through a series of convolutional and upsampling layers. However, some useful fine-grained object-specific details may be lost during bottom-up processing. To solve this problem, a series of convolutional and skipping layers~\cite{resnet} are utilized to bring the lost details back to the upsampled features. The upsampled features and lost details are combined across multiple scales as the inputs of the next module. The hourglass network takes advantage of both global and local contextual information throughout the repeated bottom-up and top-down processing. It is an ideal backbone network for deep feature extraction for visual tracking. In our approach, two hourglass networks are stacked in the backbone network.

In order to extract more powerful deep features, more hourglass modules are nested to make a deeper hourglass network. However, as the hourglass network going deeper, it becomes more complicated and not practical for real-world applications due to limited computational resources. Inspired by the recent implementation of efficient and lightweight networks, we exploit the compression method to make some critical modifications to the original hourglass network and design the architecture of a lightweight hourglass network. Instead of using the $3\times 3$ standard convolution layer to filter and consolidate input features simultaneously, we follow the depthwise separable convolutional unit described in MobileNet~\cite{mobilenet} and factorize the $3\times 3$ standard convolution into a $3\times 3$ depthwise convolution for filtering and a $1\times 1$ pointwise convolution for consolidating. In the residual block, we use a depthwise separable convolutional unit followed by a depthwise convolutional layer to extract deep features, and another pointwise convolutional layer is used to bring the original information back to the output features.

\begin{figure}[t]
\begin{center}
    \includegraphics[width=\linewidth]{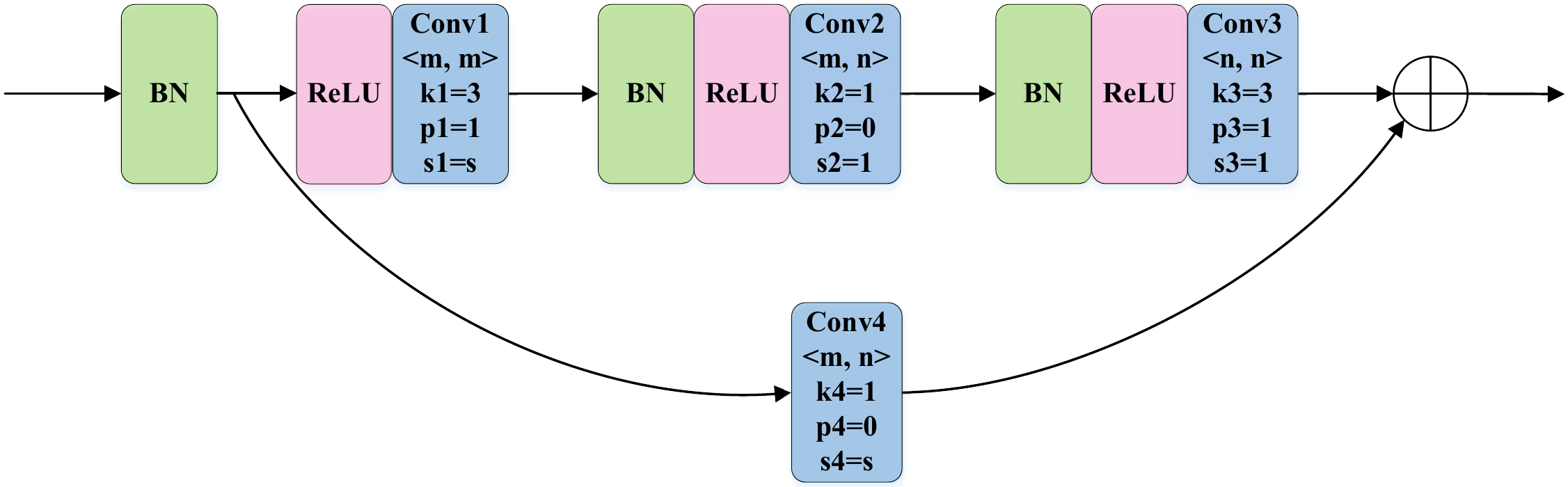}
\end{center}
\caption{Illustration of a single residual block. In the figure, \emph{BN} means the batch normalization, \emph{ReLU} means a rectified linear unit and \emph{Conv$\ast$} means a convolutional layer. The first number in the bracket indicates the input channel and the second one indicates the output channel. The parameters \emph{k$\ast$}, \emph{p$\ast$} and \emph{s$\ast$} represent the filter size, padding size and stride of the corresponding convolutional layer, respectively. The parameter \emph{s} indicates the stride of Conv1 and Conv4, which is preset according to the hourglass module.}
\label{fig:res}
\end{figure}

\begin{figure}[t]
\begin{center}
    \includegraphics[width=\linewidth]{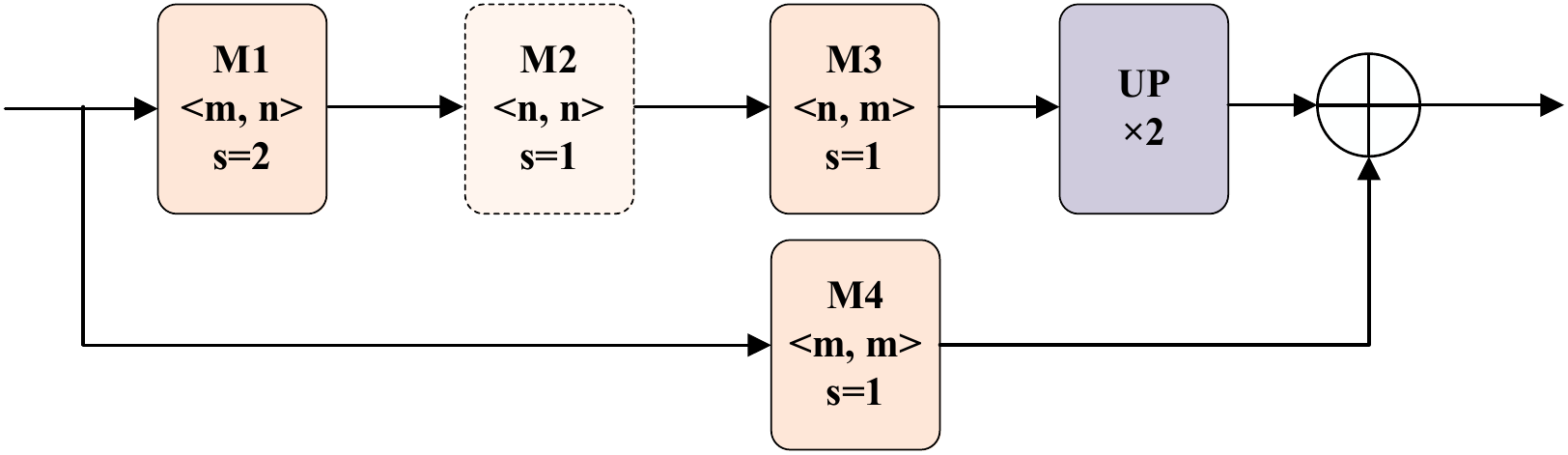}
\end{center}
\caption{Illustration of a single hourglass module. In the figure, \emph{M$\ast$} means a residual block, \emph{UP} means a upsampling layer. The first number in the bracket indicates the input channel and the second one indicates the output channel. The parameter \emph{s} represents the stride of Conv1 and Conv4 in each residual block.}
\label{fig:hour}
\end{figure}

In the hourglass module, three residual blocks are followed by an upsampling layer as the principal propagation, and another residual block works as the skipping connection. It is worth noting that the second residual block M$2$ can be nested by an hourglass module, and thus the network can do repeated bottom-up and top-down processing. Four lightweight hourglass modules are nested with output channel numbers of $(256, 512, 512, 256)$ respectively in each hourglass network. Another noteworthy modification is to reduce input feature resolution, we apply stride of 2 instead of max-pooling operation in the first residual block of each hourglass module. A simple nearest neighbor upsampling method is used to restore output feature maps to the original input resolution. After that, the upsampled features and lost details are merged directly by element-wise addition. Besides, before the first stage lightweight hourglass network, we apply an $11\times 11$ convolutional layer with stride of 2 and output channel 128 and a $5\times 5$ convolutional layer with stride 2 and output channel 256 to increase the number of input channels and reduce the input resolution. Fig.~\ref{fig:res} and Fig.~\ref{fig:hour} show implementations of our residual block and hourglass module, respectively.

\subsection{Cross-attentional module}\label{sec:33}

The backbone network as designed in Section~\ref{sec:32} is used to generate contextual representations of both the exemplar image \textbf{z} and candidate image \textbf{x}. However, those contextual representations may not be sufficient enough for tracking performance improvement because they treat every channel and region equally. Some informative channels and attentive regions of the output are more crucial for target object discrimination and localization than others, as they may involve more semantic attributes and visual patterns of different objects. In order to highlight meaningful information and to enhance the representation power of output features, an effective cross-attentional module in our approach is exploited as follows.

\begin{figure}[t]
\begin{center}
    \includegraphics[width=\linewidth]{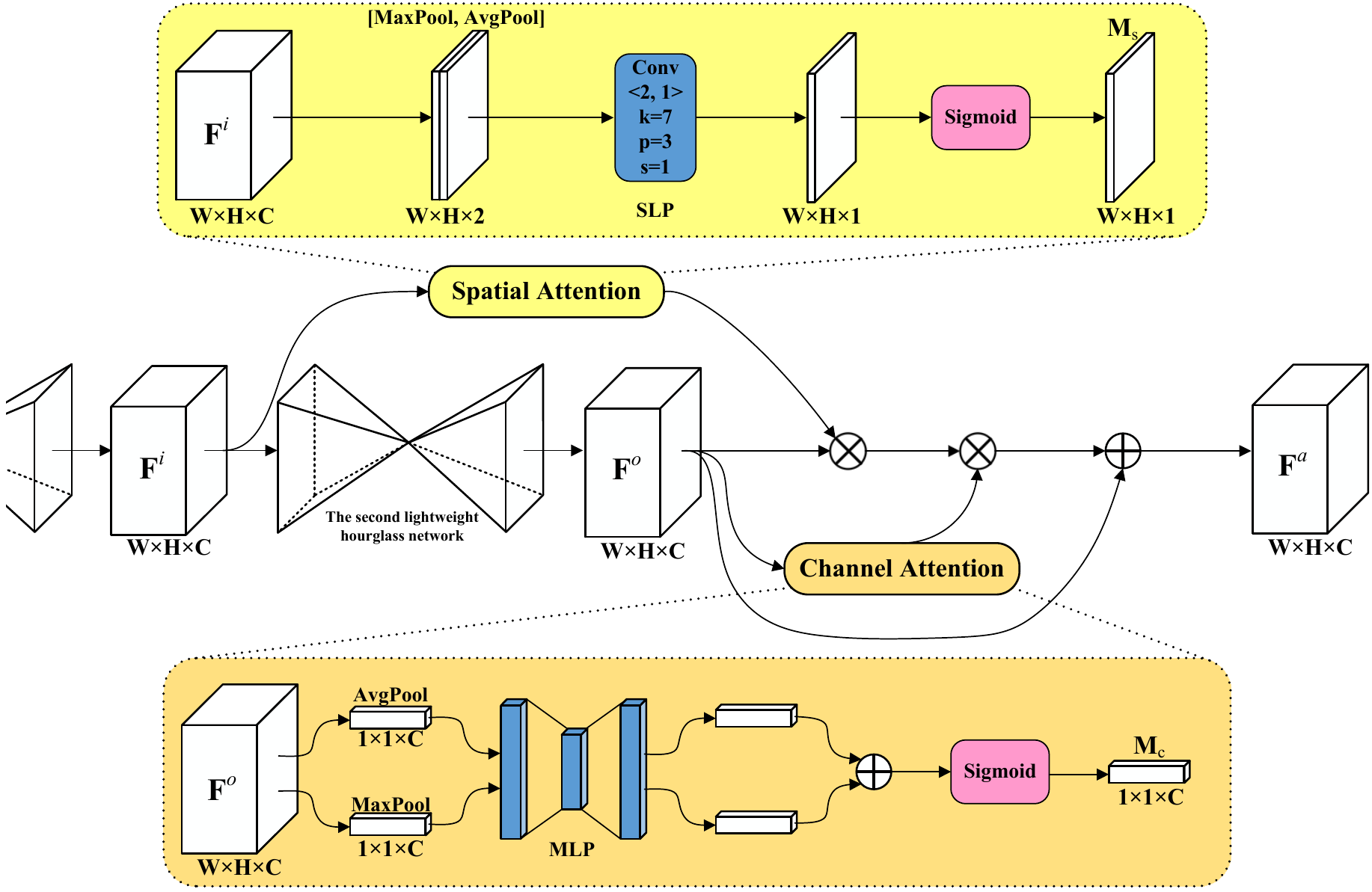}
\end{center}
\caption{The structure of the cross-attentional module. The concatenation of max-pooled and average-pooled input feature maps are supplied into a single-layer perceptron to infer a 2D spatial attention map. Both max-pooled and average-pooled output feature maps are fed into a shared multi-layer perceptron to generate a 1D channel-wise attention map. Finally, the output feature map is weighted with both channel-wise and spatial attention by broadcasting element-wise multiplication.}
\label{fig:att}
\end{figure}

Without loss of generality, the input and output feature maps of the second stage lightweight hourglass network are respectively denoted as $\mathbf{F}^i\in \mathbb{R}^{W\times H \times C}$ and $\mathbf{F}^o\in \mathbb{R}^{W\times H \times C}$, where $W$, $H$ and $C$ indicate the width, height and channel number of feature maps, respectively. Since low-level geometric representations contribute more in target object localization, and the high-level semantic information is sufficient for distinguishing the target object from the background surroundings, we utilize a single-layer perceptron on $\mathbf{F}^i$ and a multi-layer perceptron on $\mathbf{F}^o$, therefore encode those regions and channels to be emphasized or suppressed, respectively. The structure of the cross-attentional module is shown in Fig.~\ref{fig:att}.

\textbf{Spatial attention.} We utilize spatial attention to highlight informative regions that adequately represent the current target object. Given the input feature map $\mathbf{F}^i\in \mathbb{R}^{W\times H \times C}$, we first apply global average-pooling and max-pooling operations along the channel axis to construct two 2D spatial feature descriptors, denoted as $\mathbf{F}^i_{avg}\in \mathbb{R}^{W\times H \times 1}$ and $\mathbf{F}^i_{max}\in \mathbb{R}^{W\times H \times 1}$. Spatial feature descriptors are then concatenated and fed into a single-layer perceptron with a sigmoid activation, where a 2D spatial attention map $\mathbf{M}_s\in \mathbb{R}^{W\times H\times 1}$ is created over the size of $W\times H$,
\begin{equation}
    \mathbf{M}_s = \sigma\big(\verb"SLP"([\mathbf{F}^i_{avg};\mathbf{F}^i_{max}])\big)
\end{equation}
where $\sigma$ indicates the sigmoid function, and $[\mathbf{F}^i_{avg};\mathbf{F}^i_{max}]\in \mathbb{R}^{W\times H \times 2}$ denotes the concatenation of global average-pooled and max-pooled feature descriptors. $\verb"SLP"$ means the single-layer perceptron consisting of a $7\times 7$ convolutional layer with padding 3, stride 1, input channel 2 and output channel 1.

\textbf{Channel-wise attention.} The channel-wise attention is employed to select useful channels which facilitate the current tracking task adaptively. Given the output feature map $\mathbf{F}^o\in \mathbb{R}^{W\times H \times C}$, global average-pooling and max-pooling operations are applied along the spatial axis to generate two 1D channel-wise feature descriptors denoted as $\mathbf{F}^o_{avg}\in \mathbb{R}^{1\times 1 \times C}$ and $\mathbf{F}^o_{max}\in \mathbb{R}^{1\times 1 \times C}$, respectively. Different from the spatial-wise attention, a multi-layer perceptron is applied on each pooled feature descriptor to create a 1D channel attention map $\mathbf{M}_c\in \mathbb{R}^{1\times 1\times C}$ over $C$ channels,
\begin{equation}
\begin{aligned}
  \mathbf{M}_c &= \sigma\big(\verb"MLP"(\mathbf{F}^o_{avg})\oplus \verb"MLP"(\mathbf{F}^o_{max})\big) \\
  &= \sigma\Big(W_u\big(W_d(\mathbf{F}^o_{avg})\big)\oplus W_u\big(W_d(\mathbf{F}^o_{max})\big)\Big)
\end{aligned}
\end{equation}
where $\sigma$ indicates the sigmoid function, $\oplus$ denotes the element-wise addition, and $\verb"MLP"$ means the multi-layer perceptron. Moreover, the multi-layer perceptron is composed of a channel reduction layer with weight $W_d\in\mathbb{R}^{\frac{C}{r}\times C}$ and a ReLU activation, and a channel-increasing layer with weight $W_u\in\mathbb{R}^{C\times \frac{C}{r}}$ and a sigmoid activation, where $r$ (set to 4) is a channel reduction ratio~\cite{cbam} to reduce the computational burden. It is worth noting that the weights $W_d$ and $W_u$ of the multi-layer perceptron are shared for both global average-pooled and max-pooled feature descriptors.

The attention-refined feature map $\mathbf{F}^a\in \mathbb{R}^{W\times H \times C}$ weighted with both the channel-wise and spatial attention can be calculated sequentially as,
\begin{equation}
    \mathbf{F}^a=\mathbf{F}^o\oplus (\mathbf{F}^o\otimes \mathbf{M}_s\otimes \mathbf{M}_c)
\end{equation}
where $\otimes$ and $\oplus$ indicate multiplication and element-wise addition, respectively. Another noteworthy issue is that the spatial attention map $\mathbf{M}_s$ and the channel attention map $\mathbf{M}_c$ are broadcasted along the channel and the spatial axes, respectively, during element-wise multiplication. The original output information is brought back to the attention-refined feature map in the cross-attentional module.

In our approach, the cross-attentional module is only applied to the exemplar image. Before the keypoints detection process, we reduce the resolution of $\mathbf{F}^a(\mathbf{z})$ and $\mathbf{F}^o(\mathbf{x})$ by two times using $3\times 3$ convolutional layers, both with stride 2 and padding 1. For convenience, final feature maps of the exemplar image $\mathbf{z}$ and the candidate image $\mathbf{x}$ are denoted as $\mathbf{F}(\mathbf{z})$ and $\mathbf{F}(\mathbf{x})$, respectively.

\subsection{Keypoint detection}\label{sec:34}

We now show how those three keypoint detection modules work, i.e., the top-left corner point, the centroid point, and the bottom-right corner point detection module.

\textbf{Corner point detection.} The corner point detection modules are inspired by CornerNet~\cite{cornernet}, which detects the target object as a pair of bounding box corner points to eliminate the need for designing anchor boxes or multi-scale pyramids.

There are two corner point detection modules in SATIN, one for the top-left corner point detection and the other for the bottom-right corner point detection. However, because of the limitation of the receptive field and the lack of local visual pattern of corner points, it is difficult to detect the corner point outside the target object directly, as shown in Fig.~\ref{fig:heatmaps}. To address this problem, we adopt a new type of pooling layer, namely, corner pooling~\cite{cornernet}, to translate meaningful object-specific information to locations outside of the target object based on explicit prior knowledge.

\begin{figure}[t]
\begin{center}
    \includegraphics[width=\linewidth]{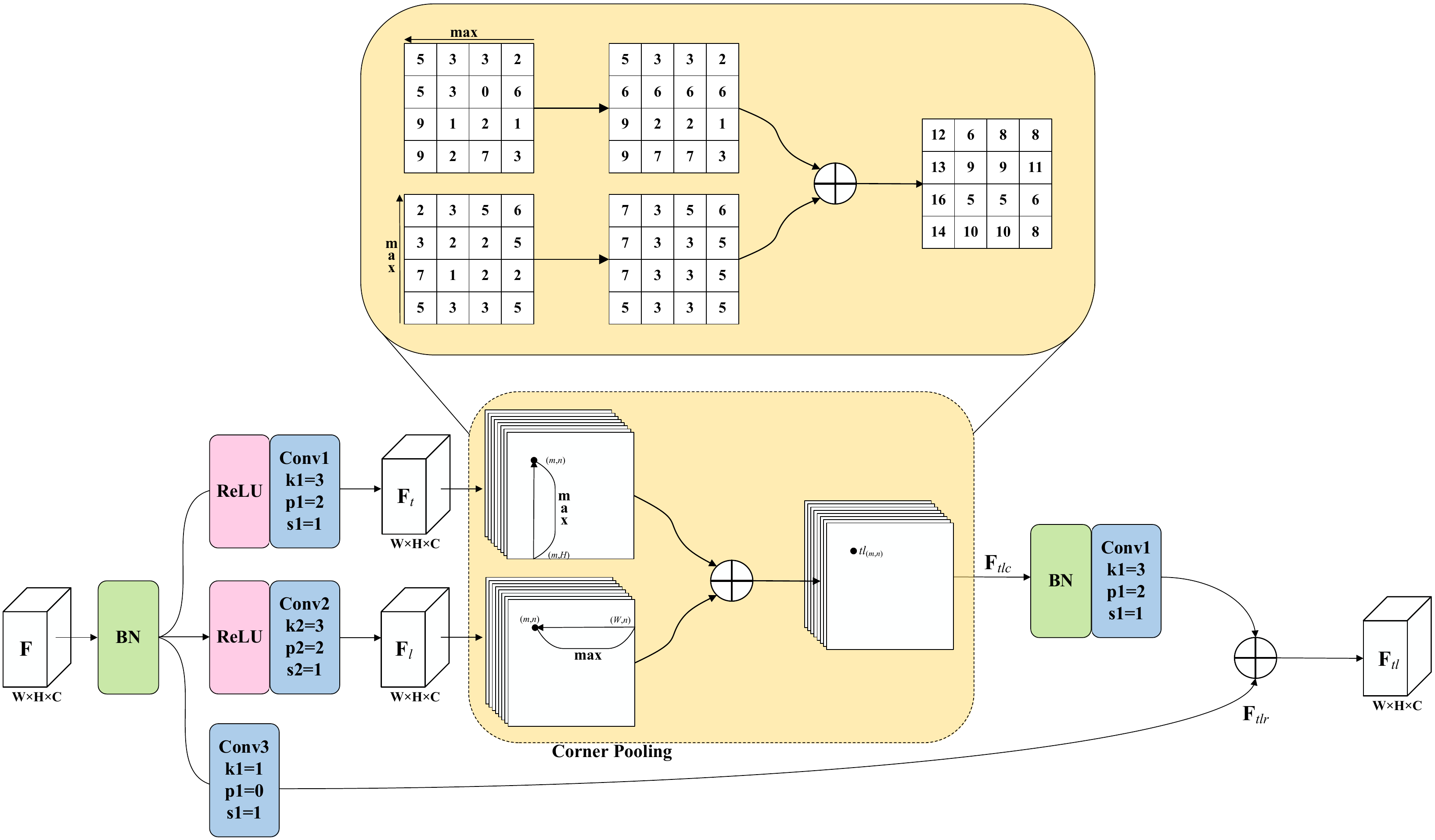}
\end{center}
\caption{An illustration of the corner-pooling operation in the top-left corner point detection module.}
\label{fig:corp}
\end{figure}

We use an example to illustrate our idea. Suppose we are going to detect the top-left corner point, as shown in Fig.~\ref{fig:corp}, the max-pooling operation is applied on the feature map horizontally from the rightmost boundary to the left, and vertically from the bottommost boundary to the top. For convenience, let $\mathbf{F}\in \mathbb{R}^{W\times H \times C}$ denote either $\mathbf{F}(\mathbf{z})$ or $\mathbf{F}(\mathbf{x})$, where $W$, $H$ and $C$ indicate the width, height and channel number of $\mathbf{F}$, respectively. The top-left corner point detection module first generates two corner-oriented feature maps $\mathbf{F}_t\in \mathbb{R}^{W\times H \times C}$ and $\mathbf{F}_l\in \mathbb{R}^{W\times H \times C}$ using two independent $3\times 3$ convolutional layers with stride 1 and padding 1, respectively. Then, we apply max-pooling operation between coordinate $(m,n)$ and $(m,H)$ on $\mathbf{F}_t$ to obtain a vertical maximum pattern $t_{(m,n)}$, where $m\in\{1,\ldots,W\}$ and $n\in\{1,\ldots,H\}$. Similarly, the same operation is applied between $(m,n)$ and $(W,n)$ on $\mathbf{F}_l$ to obtain a maximum horizontal pattern $l_{(m,n)}$. The top-left feature pattern $tl_{(m,n)}$ at location $(m,n)$ in the top-left corner-pooled feature map $\mathbf{F}_{tlc}$ is calculated by adding $t_{(m,n)}$ and $l_{(m,n)}$ together. The process is depicted as,
\begin{equation}\label{eq:5}
  \begin{aligned}
    t_{(m,n)} &= \left\{\begin{array}{ll}
                    \max(\mathbf{F}_{t_{(m,n)}},t_{(m,n+1)}) & \textrm{if $n<H$}\\
                    F_{t_{(m,H)}} & \textrm{otherwise}
                    \end{array}\right.\\
    l_{(m,n)} &= \left\{\begin{array}{ll}
                    \max(\mathbf{F}_{l_{(m,n)}},l_{(m+1,n)}) & \textrm{if $m<W$}\\
                    F_{l_{(W,n)}} & \textrm{otherwise}
                    \end{array}\right.\\
    tl_{(m,n)} &= t_{(m,n)}+l_{(m,n)}
  \end{aligned}
\end{equation}
where $\mathbf{F}_{t_{(m,n)}}$ (or $\mathbf{F}_{l_{(m,n)}}$) denote oriented feature patterns at the location $(m,n)$ in $\mathbf{F}_t$ (or $\mathbf{F}_l$). To obtain the top-left corner-specific feature map $\mathbf{F}_{tl}$, the top-left corner-pooled feature map $\mathbf{F}_{tlc}$ is then fed into a $3\times 3$ convolutional layer, and add back the residual feature map $\mathbf{F}_{tlr}$ obtained by applying a $1\times 1$ convolution to the input feature map $\mathbf{F}$.

Each corner detection module is utilized to detect only one corner point. Hence, it only outputs a single-channel heatmap for prediction and a double-channel offset map for refinement. To suppress the effect of redundant information, the top-left quarter of $\mathbf{F}_{tl}(\mathbf{z})$ is then cropped and denoted as $\big[\verb"crop"_{tl}\big(\mathbf{F}_{tl}(\mathbf{z})\big)\big]$. Therefore, two separate $3\times 3$ convolutional layers are applied to keep and double the channels of $\big[\verb"crop"_{tl}\big(\mathbf{F}_{tl}(\mathbf{z})\big)\big]$, generating $\big[\verb"crop"_{tl}\big(\mathbf{F}_{tl}(\mathbf{z})\big)\big]_{s}$ (with the same number of channels) and $\big[\verb"crop"_{tl}\big(\mathbf{F}_{tl}(\mathbf{z})\big)\big]_{o}$ (with doubled number of channels), respectively. Meanwhile, $\mathbf{F}_{tl}(\mathbf{x})$ is also divided into two branches $[\mathbf{F}_{tl}(\mathbf{x})]_{s}$ and $[\mathbf{F}_{tl}(\mathbf{x})]_{o}$ by two separate $3\times 3$ convolutional layers, but the channels are kept unchanged. The heatmap $\mathbf{S}_{lt}$ and offset $\mathbf{O}_{lt}$ are computed as,
\begin{equation}\label{eq:corner}
  \begin{aligned}
    \mathbf{S}_{lt} &= \big[\verb"crop"_{tl}\big(\mathbf{F}_{tl}(\mathbf{z})\big)\big]_{s} \ast [\mathbf{F}_{tl}(\mathbf{x})]_{s}  \\
    \mathbf{O}_{lt} &= \big[\verb"crop"_{tl}\big(\mathbf{F}_{tl}(\mathbf{z})\big)\big]_{o} \ast [\mathbf{F}_{tl}(\mathbf{x})]_{o}
  \end{aligned}
\end{equation}
where $\big[\verb"crop"_{tl}\big(\mathbf{F}_{tl}(\mathbf{z})\big)\big]_{s}$ and $\big[\verb"crop"_{tl}\big(\mathbf{F}_{tl}(\mathbf{z})\big)\big]_{o}$ are served as correlation kernels, and $\ast$ denotes the cross-correlation operation. Each score in $\mathbf{S}_{lt}$ represents the probability of being the top-left corner point at the location $(m,n)$. After upsampling $\mathbf{S}_{lt}$ to the resolution of $\mathbf{x}$, the location of the maximum score relative to the top-left corner point of the bounding box.

Due to a series of downsampling operations involved in the previous modules, the top-left corner point $(m_{lt},n_{lt})$ in $\mathbf{x}$ is mapped to the location $\big(\big\lfloor\frac{m_{lt}}{\alpha}\big\rfloor,\big\lfloor\frac{n_{lt}}{\alpha}\big\rfloor\big)$ in $\mathbf{S}_{lt}$, where $\big\lfloor\cdot\big\rfloor$ denotes the round down operation, and $\alpha$ is the downsampling factor. However, if we remap the location $\big(\big\lfloor\frac{m_{lt}}{\alpha}\big\rfloor,\big\lfloor\frac{n_{lt}}{\alpha}\big\rfloor\big)$ from $\mathbf{S}_{lt}$ to $\mathbf{x}$ by multiplying $\alpha$ directly, we may not obtain its corresponding location $(m_{lt},n_{lt})$, which significantly affects the tracking accuracy, especially when tracking small objects. To address this issue, we exploit the offset map $\mathbf{O}_{lt}$ to adjust the score locations before remapping them to the candidate image,
\begin{equation}\label{eq:offset}
    (m_{lt},\,n_{lt}) = \alpha\cdot\bigg(\Big\lfloor\frac{m_{lt}}{\alpha}\Big\rfloor + \mathbf{O}_{lt}^1,\,\Big\lfloor\frac{n_{lt}}{\alpha}\Big\rfloor + \mathbf{O}_{lt}^2\bigg)
\end{equation}
where $\mathbf{O}_{lt}^a$ indicates the $a$-th channel of $\mathbf{O}_{lt}$.

A similar procedure can be extended to the bottom-right corner point detection, but along the topmost boundary to the bottom and the leftmost boundary to the right. Two corner-oriented feature maps $\mathbf{F}_b$ and $\mathbf{F}_r$ are first generated respectively by applying two separate $3\times 3$ convolutional layers on the feature map $\mathbf{F}$. The bottom-right corner pooling operation is then applied to create the feature map $\mathbf{F}_{br}$. Both the heatmap $\mathbf{S}_{br}$ and the offset map $\mathbf{O}_{br}$ are computed with a formulation like Eq.~\ref{eq:corner}.

\textbf{Centroid detection.} The centroid point detection module is just a variant of the corner point detection module, but with the corner pooling operation removed. To detect the centroid point of the bounding box, only a quarter around the center of the exemplar feature map $\mathbf{F}(\mathbf{z})$ is cropped to reduce the effect of background information, as illustrated in Fig.~\ref{fig:arch}. After that, the heatmap $\mathbf{S}_{c}$ and the offset map $\mathbf{O}_{c}$ are created using similar correlation operations as described in Eq.~\ref{eq:corner}.

\subsection{Loss function}\label{sec:35}

During training, it is conventional to assign a binary label $y\in\{0,1\}$ for each pixel $(m,n)$ in the candidate image $\mathbf{x}$ to indicate whether it is a keypoint or not. Instead of doing this, a soft label is exploited to penalize negative locations smoothly within a radius of the ground-truth keypoint. This is because, for each keypoint, there is only one positive location, while all other locations are negative. However, false keypoints sometimes detected close to their respective positive locations, but the detected bounding box can still satisfactorily overlap the ground-truth bounding box (region of interests). To ensure those three keypoints are detected correctly, it is required that the detected bounding box must have an Intersection-over-Union (IoU) overlap higher than $0.7$ with the ground-truth bounding box, and the radius is thus determined by the size of the ground-truth bounding box. For convenience, let $(m_1,n_1)$, $(m_2,n_2)$ and $(m_3,n_3)$ indicate coordinates for the top-left corner point, the centroid point and the bottom-right corner point of the detected bounding box, respectively. $(\hat{m}_1,\hat{n}_1)$, $(\hat{m}_2,\hat{n}_2)$ and $(\hat{m}_3,\hat{n}_3)$ are those of the ground-truth bounding box. The soft labels corresponding to the pixel $(m,n)$ are defined by an unnormalized Gaussian function,
\begin{equation}\label{eq:label}
  y_{(m,n),i}=\exp\bigg(-\frac{(m-\hat{m_i})^2+(n-\hat{n_i})^2}{2\sigma^2}\bigg),\;\; i\in\{1,2,3\}
\end{equation}
where $i$ is the index of a keypoint as mentioned above, and $\sigma$ is $1/3$ of the radius. The soft labels $y_{(m,n),i}$ is centered at their corresponding ground-truth keypoints $(\hat{m}_i,\hat{n}_i)$ and smoothly reduces to 0 for the locations outside the radius.

Although we increase the number of positive locations by labeling the negative locations within a radius of the ground-truth location with soft labels, there remains an imbalance problem between the limited positive locations within the radius and a substantial amount of negative locations outside the radius. These easy negative locations will take over the majority of training losses and dominate the gradient. To solve this problem, we investigate the relationship between the soft labels $y_{(m,n),i}$ and the probability scores $s_{(m,n),i}$ corresponding to the location $(m,n)$, and propose a location-sensitive loss function $\mathcal{L}_{hm}$ of the keypoint heatmap,
\begin{equation}\label{eq:loss1}
    \mathcal{L}_{hm} = -\sum_{i=1}^{3}\sum_{m=1}^{W}\sum_{n=1}^{H}y_{(m,n),i}(1-s_{(m,n),i})\log(s_{(m,n),i})+s_{(m,n),i}(1-y_{(m,n),i})\log(1-s_{(m,n),i})
\end{equation}
where $s_{(m,n),i}$ is the score value at location $(m,n)$ associated with the $i$-th keypoint heatmap obtained in Section~\ref{sec:34}.

With the ground-truth keypoints $(\hat{m}_{i},\hat{n}_{i})$, the precision loss caused by the downsampling operations can be written as below,
\begin{equation}\label{eq:offset}
    \mathbf{\hat{O}}_{i} = \bigg(\frac{\hat{m}_{i}}{\alpha}-\Big\lfloor\frac{\hat{m}_{i}}{\alpha}\Big\rfloor,\,\frac{\hat{n}_{i}}{\alpha}-\Big\lfloor\frac{\hat{n}_{i}}{\alpha}\Big\rfloor\bigg)
    ,\;\; i\in\{1,2,3\}
\end{equation}
where $\alpha$ is the downsampling factor. Then the offset loss $\mathcal{L}_{os}$ can be formulated as,
\begin{equation}\label{eq:loss2}
  \mathcal{L}_{os} = \sum_{i=1}^{3}\sum_{k=1}^{2}smooth_{L1}\Big(\mathbf{O}_i^k-\mathbf{\hat{O}}_i^k\Big)
\end{equation}
where $\mathbf{O}_i$ is the offset map predicted in Section~\ref{sec:34}, and $smooth_{L1}$ is a robust loss function~\cite{fastrcnn} defined as,
\begin{equation}\label{eq:smooth}
  smooth_{L1}(x)=\left\{\begin{array}{ll}
                    0.5x^2 & \textrm{if $|x|<1$}\\
                    |x|-0.5 & \textrm{otherwise}
                    \end{array}\right.\\
\end{equation}

Since the top-left corner point and the bottom-right corner point should be symmetric around the centroid point, the relative distance between the top-left corner point and the centroid point must be equal to that between the bottom-right corner point and the centroid point. Hence, the symmetry loss is defined as,
\begin{equation}\label{eq:loss3}
  \mathcal{L}_{st}=\big((m_2-m_1)-(m_3-m_2)\big)+\big((n_2-n_1)-(n_3-n_2)\big)
\end{equation}

Finally, our SATIN is trained by minimizing the combination loss $\mathcal{L}$ of the heatmap loss $\mathcal{L}_{hm}$, the offset loss $\mathcal{L}_{os}$, and the symmetry loss $\mathcal{L}_{st}$ as,
\begin{equation}\label{eq:loss}
  \mathcal{L} = \mathcal{L}_{hm}+\lambda_1\mathcal{L}_{os}+\lambda_2\mathcal{L}_{st}
\end{equation}
where $\lambda_1$ and $\lambda_2$ are trade-off factors to balance the three loss.

\section{Experiments}\label{sec:4}

In this section, implementation details are introduced first. Then extensive experiments on OTB and VOT benchmark datasets are conducted to evaluate our proposed approach. Finally, ablation studies are carried out on OTB benchmark datasets to investigate how each component contributes to improving performance.

\subsection{Implementation details}

We apply stochastic gradient descent (SGD) with the momentum of $0.9$ and the weight decay of $0.005$ to train SATIN offline from scratch using Image-VID~\cite{imagenet} and YouTube-VOS~\cite{youtube}. Totally $50$ epochs are performed with learning rate decreased in a logarithmic manner from $10^{-3}$ to $10^{-5}$. Frame pairs are collected with interval less than $100$ from the same video sequence, and exemplar images and candidate images are then cropped from frames pairs around their corresponding ground-truth bounding box with $2\times$ and $4\times$ padding respectively. Random translations are performed up to $\pm12$ pixels, and the range of rescaling varies from $2^{-1/4}$ to $2^{1/4}$. If the cropped image extends exceed the size of the frame, the missing portion is filled with the mean background RGB value. After that, exemplar and candidate images are resized to the size of $127\times 127 \times 3$ and $255\times 255\times 3$, respectively. The trade-off factors in Eq.~\ref{eq:loss} are set as $\lambda_1=1$ and $\lambda_2=10$. Before we train the full approach, the backbone network is pre-trained individually on ImageNet-DET~\cite{imagenet} and COCO~\cite{coco} for warm-up to enhance both generalization and representation capabilities. Similar to~\cite{hourglass}and~\cite{cornernet}, we add intermediate supervision in training without adding the intermediate predictions back to the network.

\begin{figure}[t]
\begin{center}
    \includegraphics[width=\linewidth]{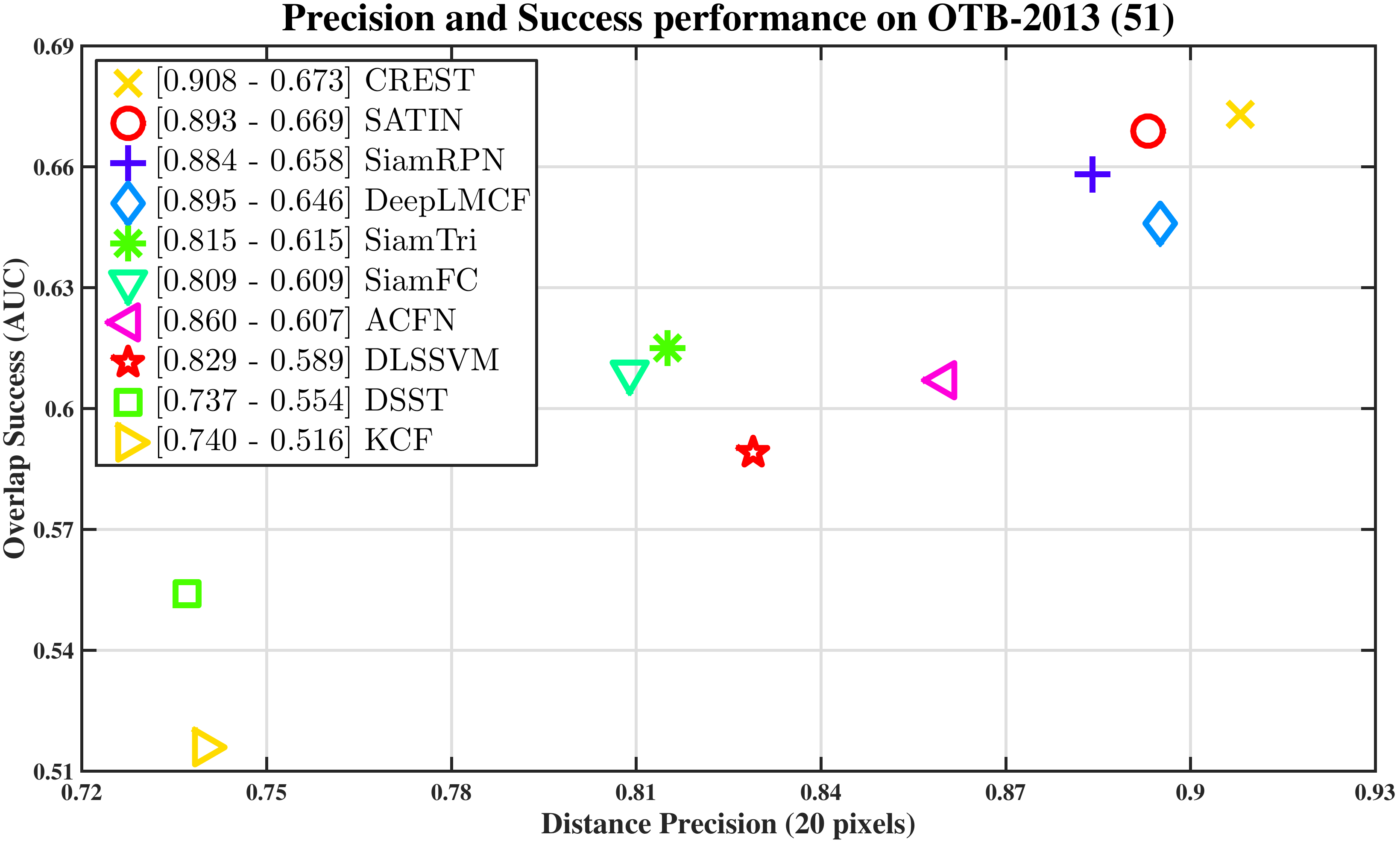}
\end{center}
\caption{Success and precision performance on the OTB-2013 benchmark~\cite{otb2013}. The first number in the legend indicates the AUC score while the second denotes DP score at error threshold of 20 pixels. The best trackers are located at the top-right corner.}
\label{fig:otb2013}
\end{figure}

\begin{figure}[t]
\begin{center}
    \includegraphics[width=\linewidth]{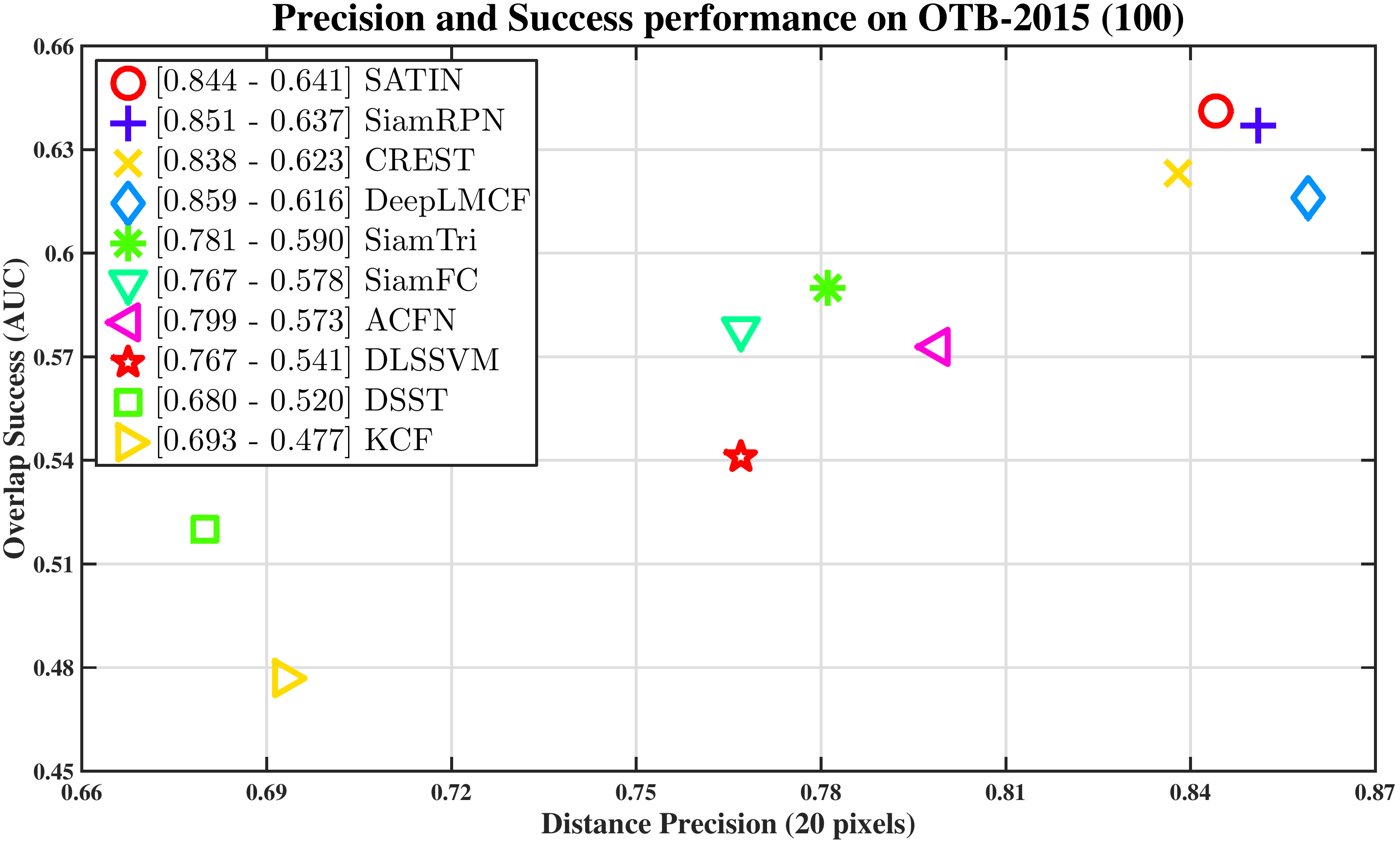}
\end{center}
\caption{Success and precision performance on the OTB-2015 benchmark~\cite{otb2015}. The first number in the legend indicates the AUC score while the second denotes DP score at error threshold of 20 pixels. The best trackers are located at the top-right corner.}
\label{fig:otb2015}
\end{figure}

Our approach is implemented using MXNet~\cite{mxnet}, trained on an Amazon EC2 instance with $16$ Intel$^\circledR$ Xeon$^\circledR$ E5-2686 v4 @ 2.3GHz CPU with 732GB RAM, and 16 NVIDIA$^\circledR$ Tesla$^\circledR$ K80 GPU with $192$GB VRAM. Experiments are conducted on an Amazon EC2 instance with an Intel$^\circledR$ Xeon$^\circledR$ E5-2686 v4 @ 2.3GHz CPU with 61GB RAM, and an NVIDIA$^\circledR$ Tesla$^\circledR$ K80 GPU with 12GB VRAM. All the parameters are fixed across experiments and datasets. The average tracking speed of our proposed tracker is $27$ frames per second (fps), which exceeds the real-time bound of $15$ fps~\cite{survey2011} by a large margin.

\begin{figure}[t]
\begin{center}
    \includegraphics[width=\linewidth]{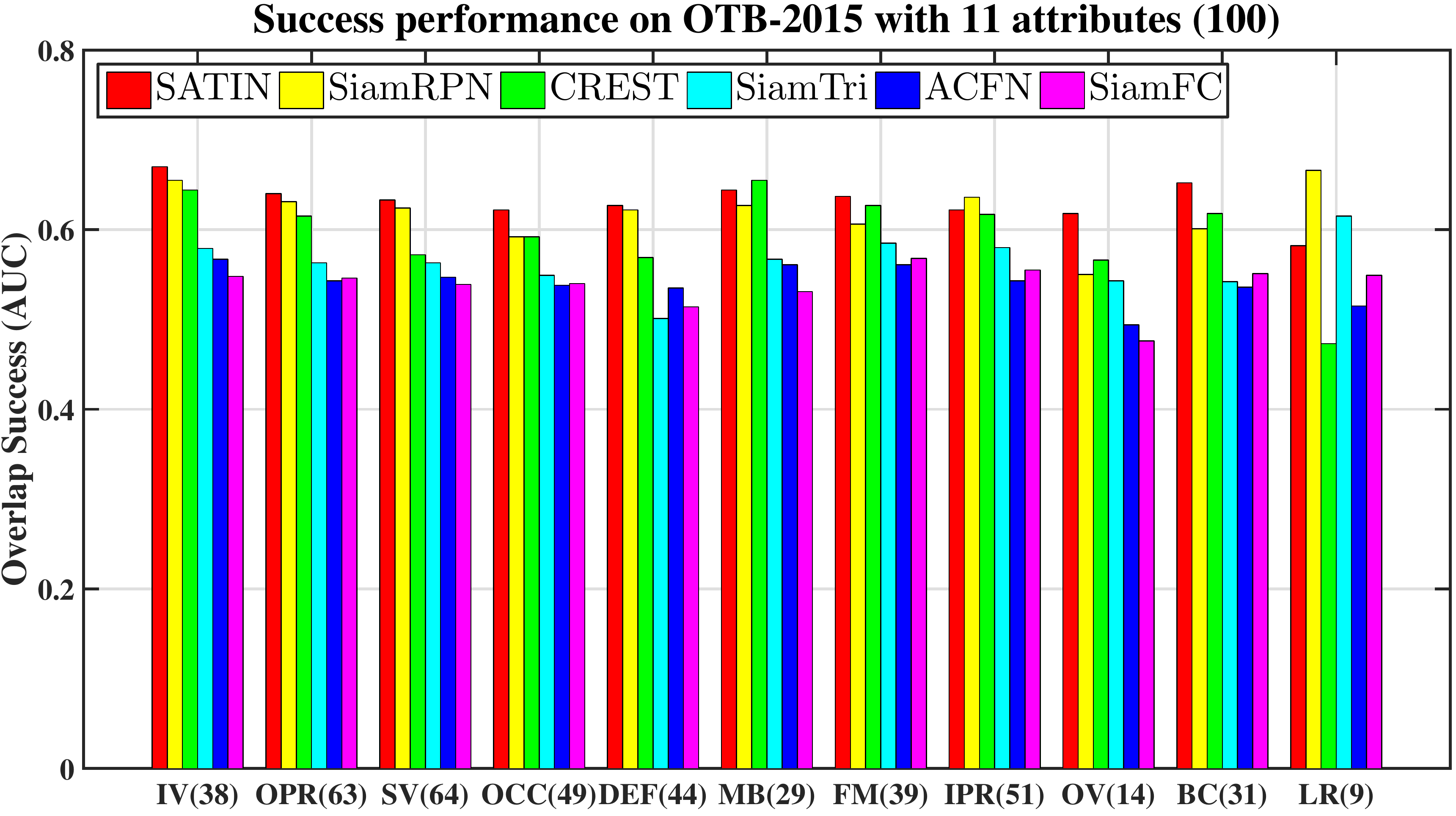}
\end{center}
\caption{Success performance on the OTB-2015 benchmark with 11 attributes~\cite{otb2015}. Best viewed in color.}
\label{fig:attauc}
\end{figure}

\subsection{Experiments on OTB}

We compare SATIN with nine state-of-the-art trackers, including ACFN~\cite{acfn}, CREST~\cite{crest}, DSST~\cite{dsst}, SiamTri~\cite{siamtri}, SiamRPN~\cite{siamrpn}, SiamFC~\cite{siamfc}, DeepLMCF~\cite{lmcf}, DLSSVM~\cite{dlssvm} and KCF~\cite{kcf}, on the OTB benchmark datasets~\cite{otb2013,otb2015}. Among the participants, we treat SiamFC and SiamRPN as our baseline. Two widely-used metrics: distance precision (DP) and overlap success (OS) are employed to evaluate tracking performance. DP is the percentage of frames in the video where the Euclidean distance between the detected object location and the ground-truth object location is smaller than a fixed threshold of $20$ pixels. OS is the percentage of frames in the video where the overlap ratio between the detected bounding box and the ground-truth bounding box exceeds a preset threshold of $0.5$. All the trackers are initialized with the ground-truth bounding box in the first frame. For better performance measurement, we utilize the area-under-curve (AUC) to present the OS score within the threshold range of $[0,1]$~\cite{otb2013}.

Fig.~\ref{fig:otb2013} illustrates the \emph{precision and success performance} under one-pass evaluation (OPE) over the $51$ fully-annotated video sequences in the OTB-2013 benchmark dataset~\cite{otb2013}. SATIN achieves the second-best AUC score of $66.9\%$ and the third-best DP score of $89.3\%$, respectively. The residual tracker CREST~\cite{crest} performs favorably against all compared trackers, slightly outperforming SATIN by gains of $0.4\%$ and $1.5\%$ in AUC and DP scores, respectively. Although both DeepLMCF~\cite{lmcf} and DLSSVM~\cite{dlssvm} employ support vector machines (SVM) as classifiers to distinguish the target object from the surrounding background, their performance is still left far behind by SATIN. Compared with the pioneering Siamese approach SiamFC~\cite{siamfc} with an AUC score of $60.9\%$ and a DP score of $80.9\%$, SATIN obtains significant improvements by $6.0\%$ and $8.4\%$, respectively. SiamRPN~\cite{siamrpn} exploits the Siamese region proposal network to get more accurate bounding boxes. It provides an AUC score of $65.8\%$ and a DP score of $88.4\%$. Compared with SiamRPN, SATIN surpasses it by absolute gains of $1.6\%$ and $1.1\%$ in AUC and DP scores, respectively. Overall, the evaluation results on the OTB-2013 benchmark dataset demonstrate that our proposed SATIN tracker performs well against state-of-the-art trackers over the $51$ challenging video sequences.
\begin{figure}[t]
\begin{center}
    \includegraphics[width=\linewidth]{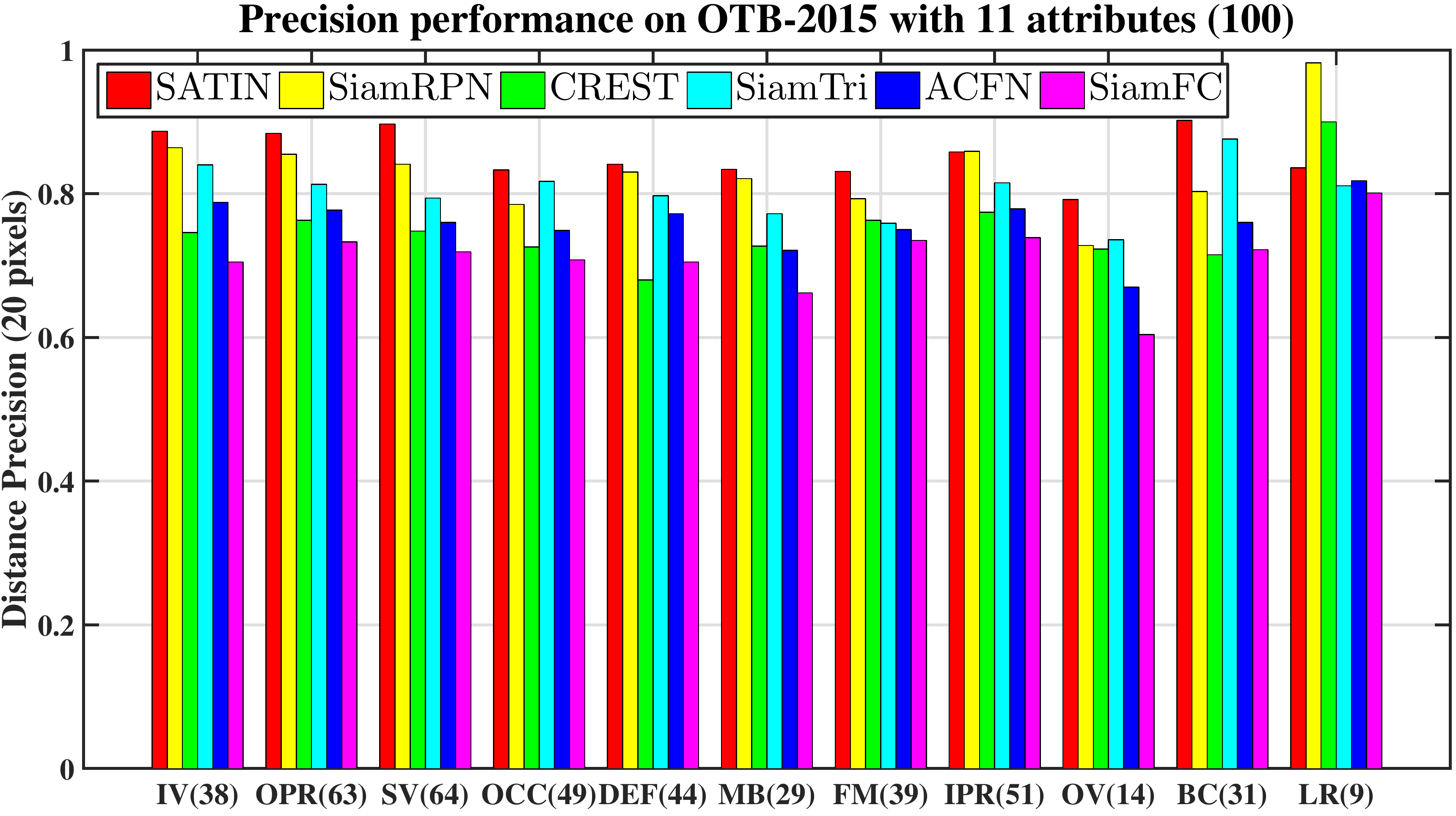}
\end{center}
\caption{Precision performance on the OTB-2015 benchmark with 11 attributes~\cite{otb2015}. Best viewed in color.}
\label{fig:atterror}
\end{figure}

Fig.~\ref{fig:otb2015} illustrates the \emph{precision and success performance} under OPE over the $100$ fully-annotated video sequences in the OTB-2015 benchmark dataset~\cite{otb2015}. SATIN achieves the best AUC score and the third-best DP score, respectively. Compared with SiamFC, SATIN improves the AUC score from $57.8\%$ to $64.1\%$, and the DP score from $76.7\%$ to $84.4\%$. CREST has provided the best performance on the OTB-2013 benchmark dataset, but SATIN outperforms it by amazing improvements of $1.8\%$ and $0.6\%$ on the AUC and DP scores after extending the dataset to $100$ video sequences. In particular, SATIN is superior to the baseline SiamRPN tracker in the AUC score with an absolute gain of $1.0\%$. However, SiamRPN achieves a DP score of $85.1\%$, which provides an absolute gain of $0.7\%$ compared to SATIN. Our approach performs better than the triplet loss tracker SiamTri and gets $5.1\%$ and $6.3\%$ absolute gains in terms of AUC and DP scores. Although DeepLMCF obtains the best DP performance of $85.9\%$, its AUC score of $61.6\%$ is much inferior to SATIN. Besides, SATIN achieves absolute AUC/DP gains of $16.4\%$/$15.1\%$, $12.1\%$/$16.4\%$, $10.0\%$/$7.7\%$ and $6.8\%$/$4.5\%$ over KCF, DSST, DLSSVM and ACFN, respectively. The excellent tracking performance of SATIN can be primarily attributed to two aspects. First, the specially designed backbone network and the cross-attentional module are utilized to enhance the representation and generalization capabilities of SATIN. Second, SATIN tracks target objects by regressing keypoints of the bounding boxes to soft labels instead of generating redundant region proposals or designing complicated scale pyramids, which are robust to target object appearance variations and scale changes.

\begin{figure}[t]
\begin{center}
    \begin{minipage}[b]{\linewidth}\centerline{\includegraphics[width=\textwidth]{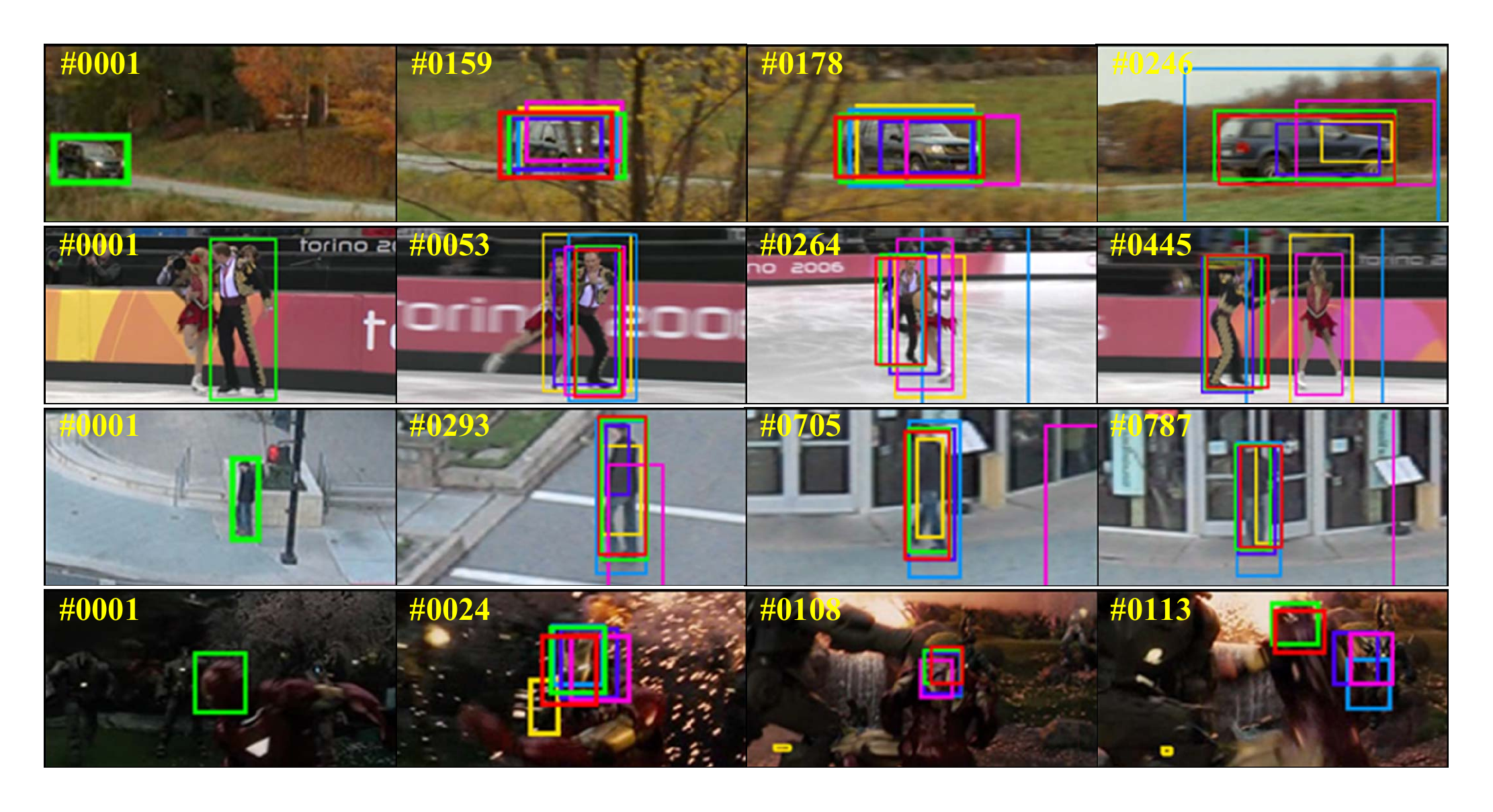}}\end{minipage}
    \vfill
    \begin{minipage}[b]{0.85\linewidth}\centerline{\includegraphics[width=\textwidth]{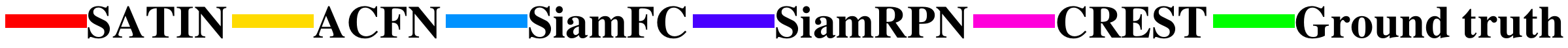}}\end{minipage}
\end{center}
\caption{Example tracking results of STAIN with the comparison to state-of-the-are trackers including ACFN~\cite{acfn}, SiamFC~\cite{siamfc}, SiamRPN~\cite{siamrpn} and CREST~\cite{crest} on four challenging sequences (from top to down: \emph{carScale}, \emph{Skating2-2}, \emph{Human6} and \emph{ironman}). The ground-truth are illustrated with green bounding boxes. Best viewed in color.}
\label{fig:otb}
\end{figure}

For comprehensive evaluation and analysis on the effectiveness of our approach, we compare SATIN with five state-of-the-art trackers, including SiamRPN~\cite{siamrpn}, CREST~\cite{crest}, SiamTri~\cite{siamtri}, ACFN~\cite{acfn}, and SiamFC~\cite{siamfc} in different attributes on the OTB-2015 benchmark. Each sequence in the OTB-2015 benchmark dataset is categorized with 11 attributes. Among all the 100 fully-annotated video sequences, there are 38 sequences with illumination variation (IV), 63 sequences with out-of-plane rotation (OPR), 64 sequences with scale variation (SV), 49 sequences with occlusion (OCC), 44 sequences with deformation (DEF), 29 sequences with motion blur (MB), 39 sequences with fast motion (FM), 51 sequences with in-plane rotation (IPR), 14 sequences with out of view (OV), 31 sequences with background clutter (BC) and 9 sequences with low resolution (LR). The results of overlap success and distance precision are shown in Fig.~\ref{fig:attauc} and Fig.~\ref{fig:atterror}, respectively. It is clear that the proposed SATIN obtains the best performance under $8$ of $11$ attributes. However, SATIN is slightly worse than SiamRPN on the attributes of in-plane rotation and low resolution. The main reason is that our backbone network may lose some fine-grained information through the repeated bottom-up and top-down framework. Besides, SATIN performs more accurate and robust under occlusion and background clutter with the help of attentional feature refinement, which can emphasize the informative feature regions and channels and reduce the effect of background noise to facilitate tracking tasks. Fig.~\ref{fig:otb} shows comparison results of SATIN with ACFN, SiamFC, SiamRPN, and CREST on OTB benchmark datasets. SATIN performs well on all the presented video sequences including \emph{carScale} (with attributes of SV, OCC, FM, IPR, and OPR), \emph{Skating2-2} (with attributes of SV, OCC, DEF, FM, and OPR), \emph{Human6} (with attributes of SV, OCC, DEF, FM, OPR, and OV) and \emph{ironman} (with attributes of IV, SV, OCC, MB, FM, IPR, OPR, OV, BC, and LR). All the results clearly demonstrate the excellent performance of SATIN.

\begin{figure}[t]
\begin{center}
    \includegraphics[width=\linewidth]{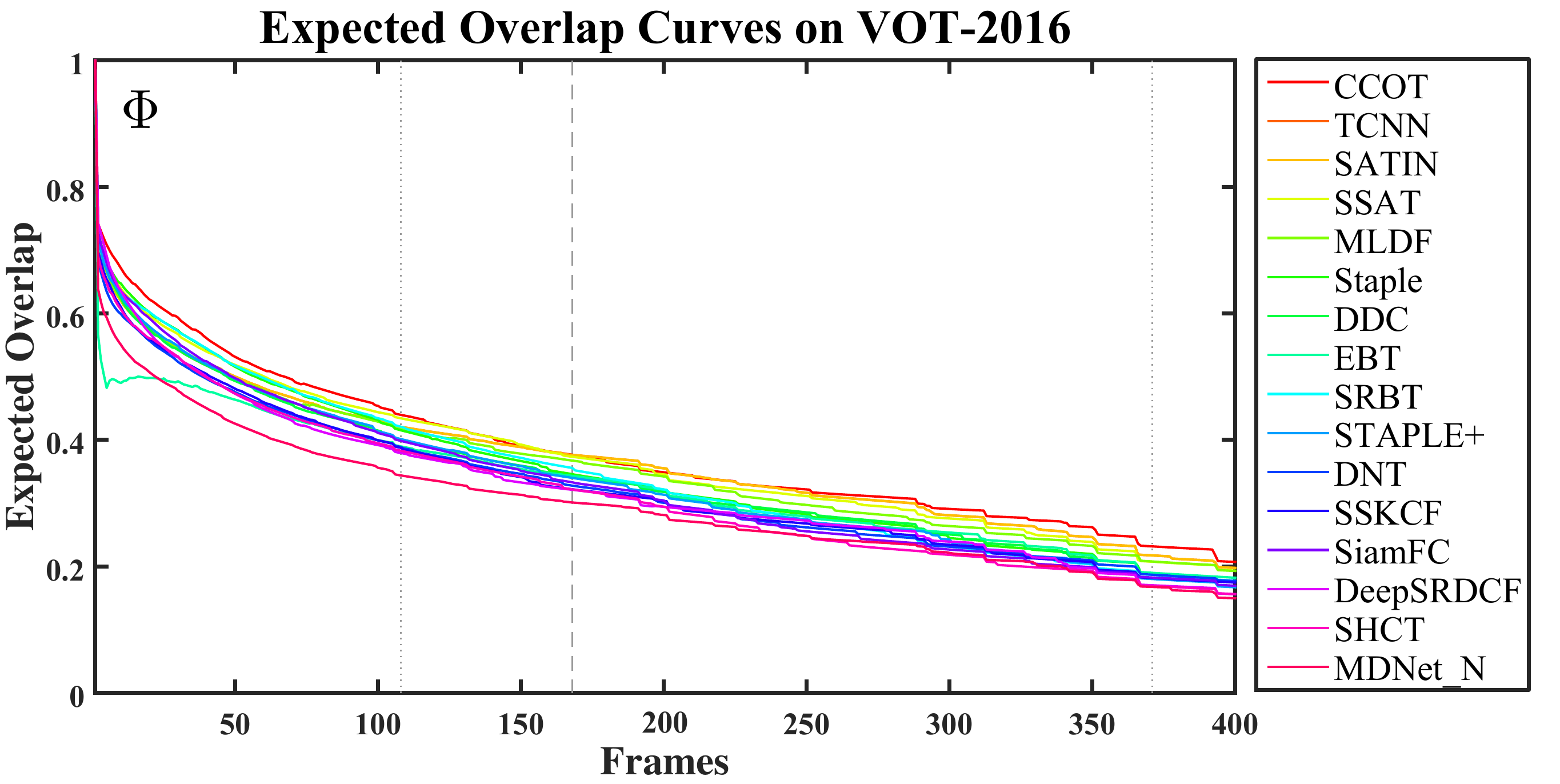}
\end{center}
\caption{Expected average overlap curve on the VOT-2016 challenge dataset~\cite{vot2016}. For presentation clarity, only the top 16 trackers with respect to the EAO score are shown in the plot.}
\label{fig:vot2016graph}
\end{figure}

\begin{figure}[t]
\begin{center}
    \includegraphics[width=\linewidth]{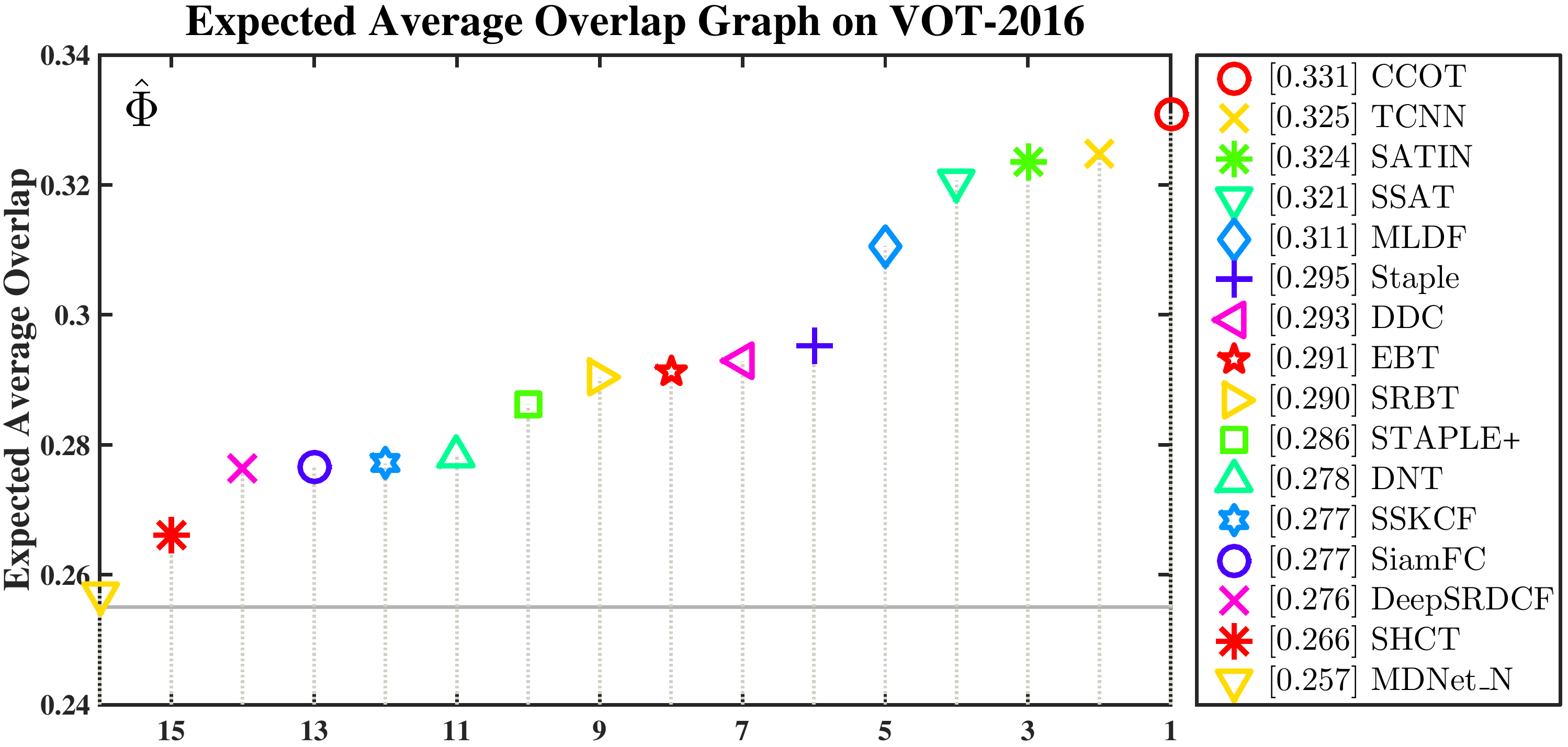}
\end{center}
\caption{Expected average overlap graph with trackers ranked on the VOT-2016 challenge dataset~\cite{vot2016}. For presentation clarity, only the top 16 trackers with respect to the EAO score are shown in the plot. The better trackers are located at the right.}
\label{fig:vot2016}
\end{figure}

\begin{table}[t]
\begin{center}
\caption{Comparisons between STAIN and 15 state-of-the-art trackers on the VOT-2016 challenge dataset. The results are presented in terms of expected average overlap (EAO), accuracy (A), robustness (R), and equivalent filter operations (EFO). The first, second and third best values are highlighted in \textbf{\textcolor[rgb]{1.00,0.00,0.00}{red}}, \textbf{\textcolor[rgb]{0.00,0.00,1.00}{blue}} and \textbf{\textcolor[rgb]{0.00,1.00,0.00}{green}} fonts, respectively.}
\label{table:vot2016}
\begin{tabular}{p{2.5cm}p{2cm}<{\centering}p{2cm}<{\centering}p{2cm}<{\centering}p{2cm}<{\centering}}
\toprule
\textbf{Tracker} & \textbf{EAO} $\uparrow$ & \textbf{A} $\uparrow$ & \textbf{R} $\downarrow$ & \textbf{EFO} $\uparrow$ \\
\midrule
CCOT~\cite{ccot} & \textbf{\textcolor[rgb]{1.00,0.00,0.00}{0.331}} & 0.539 & \textbf{\textcolor[rgb]{0.00,0.00,1.00}{0.850}} & 0.507  \\
TCNN~\cite{tcnn} & \textbf{\textcolor[rgb]{0.00,0.00,1.00}{0.325}} & 0.554 & 0.959 & 1.049  \\
SSAT~\cite{vot2016} & 0.321 & \textbf{\textcolor[rgb]{1.00,0.00,0.00}{0.577}} & 1.041 & 0.475  \\
MLDF~\cite{vot2016} & 0.311 & 0.490 & \textbf{\textcolor[rgb]{1.00,0.00,0.00}{0.833}} & 1.483  \\
Staple~\cite{staple} & 0.295 & 0.544 & 1.350 & 11.114  \\
DDC~\cite{vot2016} & 0.293 & 0.541 & 1.233 & 0.198  \\
EBT~\cite{ebtt} & 0.291 & 0.465 & \textbf{\textcolor[rgb]{0.00,1.00,0.00}{0.900}} & 3.011  \\
SRBT~\cite{vot2016} & 0.290 & 0.496 & 1.250 & 3.688  \\
STAPLEp~\cite{vot2016} & 0.286 & \textbf{\textcolor[rgb]{0.00,1.00,0.00}{0.557}} & 1.317 & \textbf{\textcolor[rgb]{1.00,0.00,0.00}{44.765}}  \\
DNT~\cite{dnt} & 0.278 & 0.515 & 1.176 & 1.127  \\
SSKCF~\cite{vot2016} & 0.277 & 0.547 & 1.333 & \textbf{\textcolor[rgb]{0.00,0.00,1.00}{29.153}}  \\
SiamFC~\cite{siamfc} & 0.277 & 0.549 & 1.367 & 5.444  \\
DeepSRDCF~\cite{deepsrdcf} & 0.276 & 0.528 & 1.167 & 0.380  \\
SHCT~\cite{vot2016} & 0.266 & 0.547 & 1.417 & 0.711  \\
MDNet\_N~\cite{mdnet} & 0.257 & 0.541 & 1.204 & 0.534  \\
\midrule
SATIN & \textbf{\textcolor[rgb]{0.00,1.00,0.00}{0.324}} & \textbf{\textcolor[rgb]{0.00,0.00,1.00}{0.561}} & 1.136 & \textbf{\textcolor[rgb]{0.00,1.00,0.00}{11.688}} \\
\bottomrule
\end{tabular}
\end{center}
\end{table}

\subsection{Experiments on VOT}

The VOT challenge~\cite{vot} is the most significant annual competition in the field of visual tracking. Each of VOT-2016, VOT-2017, and VOT-2018 benchmark datasets~\cite{vot2016,vot2017,vot2018} contains 60 video sequences. We use them to evaluate our proposed approach. In the experiments, the performance is assessed in three metrics: Accuracy, Robustness, and Expected Average Overlap (EAO). Accuracy is defined as the average overlap between the detected bounding box and the ground truth bounding box during successful tracking periods. Robustness is defined as how many times tracking failures occur. EAO represents the average overlap with no re-initialization following a failure. For fair comparisons, we use the original results provided by the VOT challenge committee~\footnote{\url{http://www.votchallenge.net/}}.

\begin{figure}[t]
\begin{center}
    \includegraphics[width=\linewidth]{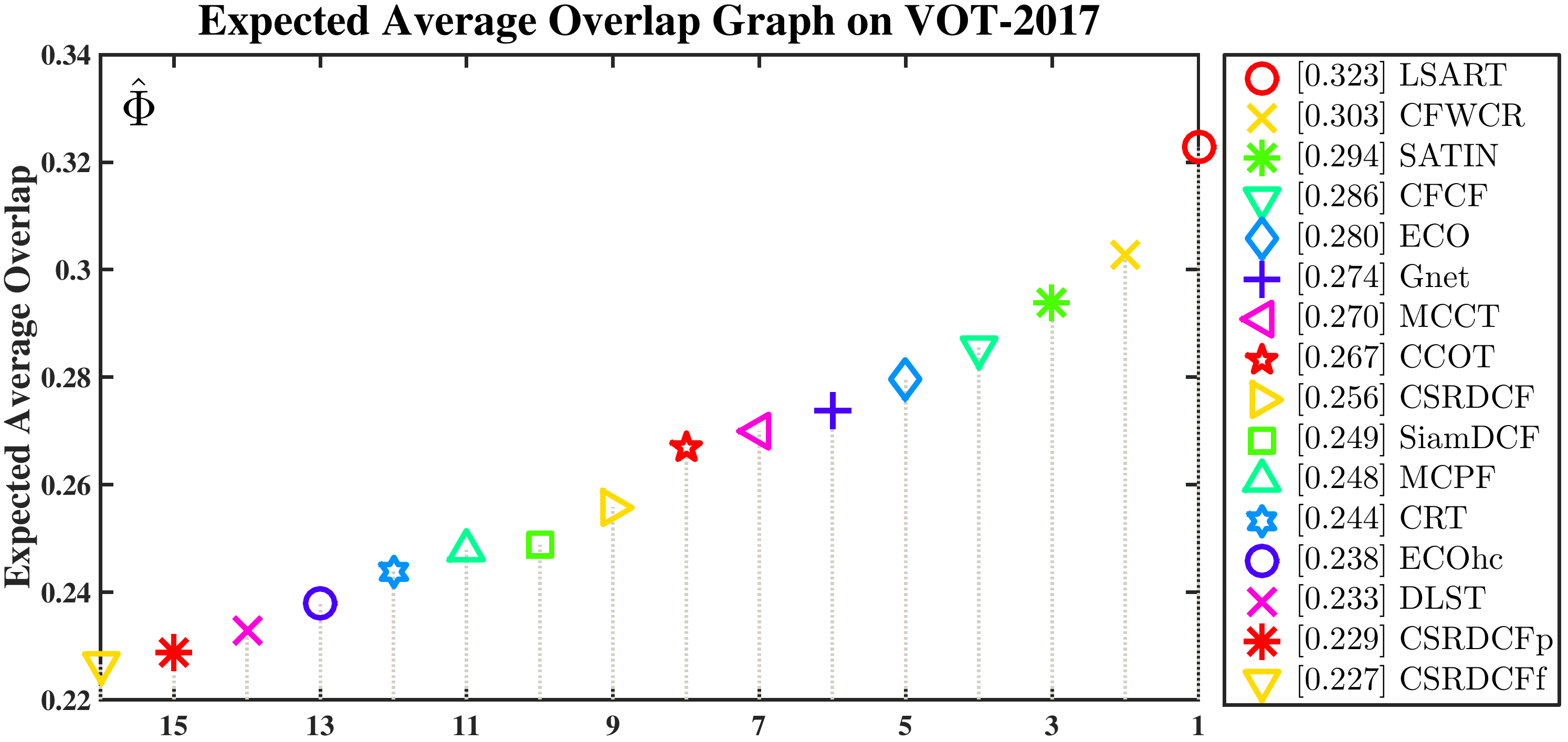}
\end{center}
\caption{Expected average overlap graph with trackers ranked on the VOT-2017 challenge dataset~\cite{vot2017}. For presentation clarity, only the top 16 trackers with respect to the EAO score are shown in the plot. The better trackers are located at the right.}
\label{fig:vot2017}
\end{figure}

We compared SATIN with $70$ other state-of-the-art trackers on the VOT-2016 challenge dataset, and report EAO curves and graph of the top $16$ participants in Fig.~\ref{fig:vot2016}. Among all the trackers, CCOT~\cite{ccot} achieves the best EAO score of $0.331$, which exploits continuous convolution operators on multi-level feature maps. It is clear that SATIN is superior to most of the state-of-the-art, and obtains the third-best EAO score of $0.324$, which is only $0.001$ lower than the second-best tracker TCNN~\cite{tcnn}. However, there is still a gap compared with the top performer CCOT~\cite{ccot}. Compared with SiamFC~\cite{siamfc} and DeepSRDCF~\cite{srdcf}, SATIN achieves absolute gains of $4.7\%$ and $4.8\%$ in EAO, respectively. Table~\ref{table:vot2016} reports the detailed comparison of SATIN with other tops 15 approaches in VOT-2016~\cite{vot2016} with respect to EAO, Accuracy, Robustness, and EFO is used to measure tracking speed. Among these 16 state-of-the-art trackers, SATIN achieves the second-best accuracy score of $0.561$ and the sixth-best robustness score of $1.136$, respectively. Compared with the baseline SiamFC~\cite{siamfc} which has an EAO score of $0.277$, SATIN substantially obtains a substantial gain of $17.0\%$ in EAO and demonstrates the superiority in terms of accuracy and robustness. Besides, compared to the top-performing tracker, i.e., CCOT, SATIN is $20\times$ faster than it with respect to the EFO score. All the above experiments show that our proposed approach performs well compared with other state-of-the-art trackers in the VOT-2016 challenge dataset.

For the evaluation on the VOT-2017 challenge dataset, Fig.~\ref{fig:vot2017} reports the EAO score of ours in comparison with $15$ other state-of-the-art trackers. It is worth noting that although the VOT-2017 challenge dataset replaces ten videos in the VOT-2016 challenge dataset with $10$ complicated sequences. Compared with the top-ranked tracker in the VOT-2016 challenge, i.e., CCOT, the proposed approach improves the EAO score from $0.267$ to $0.291$, obtaining an absolute gain of $2.4\%$. Compared with SiamDCF, which performs correlation analysis on multi-level feature maps, SATIN obtains a significant EAO gain of $4.2\%$. This is mainly attributed to our tracker takes advantages of the benefits of both global and local contextual representations across multiple scales.%

\begin{figure}[t]
\begin{center}
    \includegraphics[width=\linewidth]{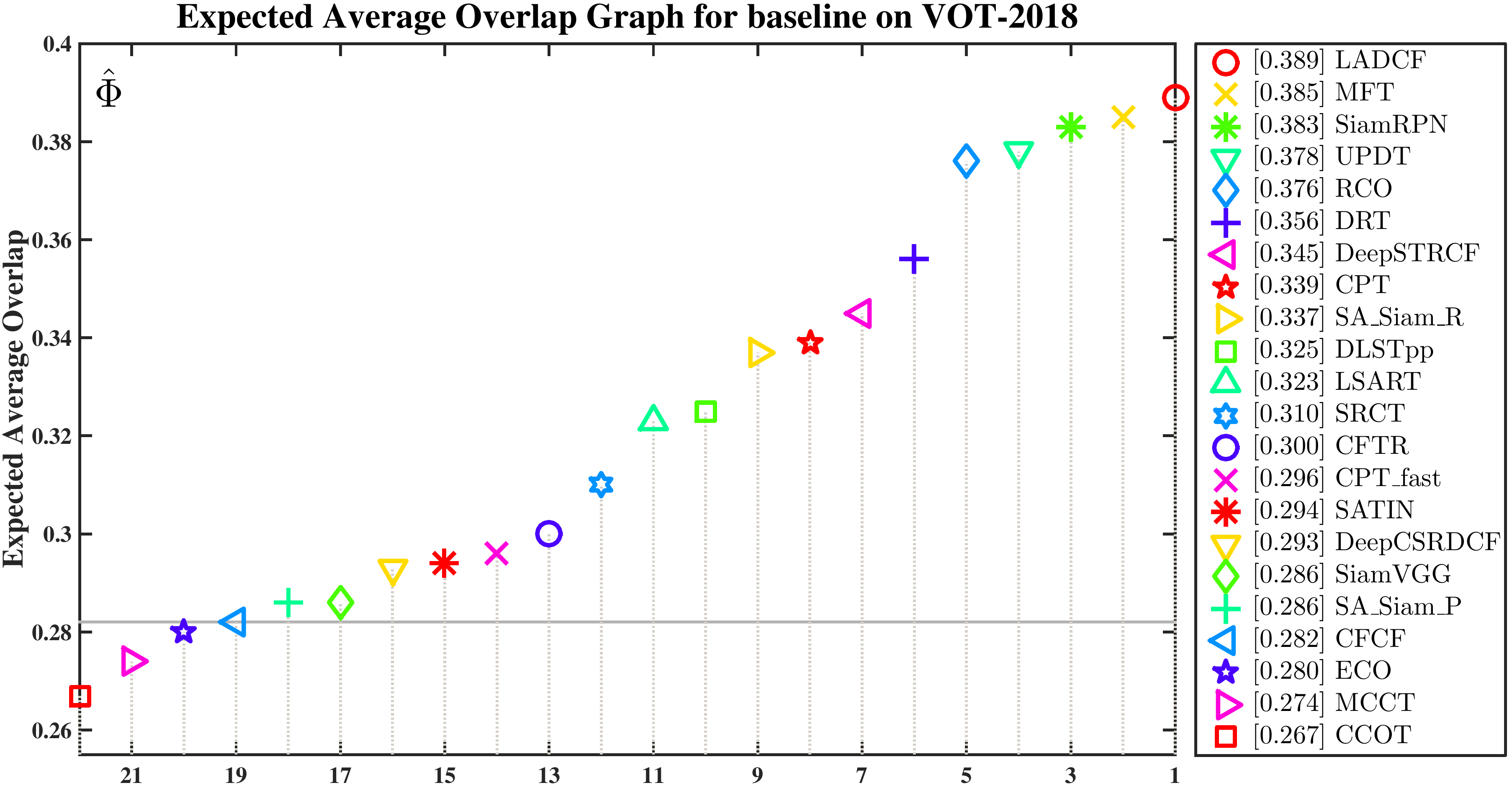}
\end{center}
\caption{Expected average overlap graph for the baseline experiment on the VOT-2018 challenge dataset~\cite{vot2018}. For presentation clarity, only the top 22 trackers with respect to the EAO score are shown in the plot. The better trackers are located at the right.}
\label{fig:vot2018base}
\end{figure}

\begin{figure}[t]
\begin{center}
    \includegraphics[width=\linewidth]{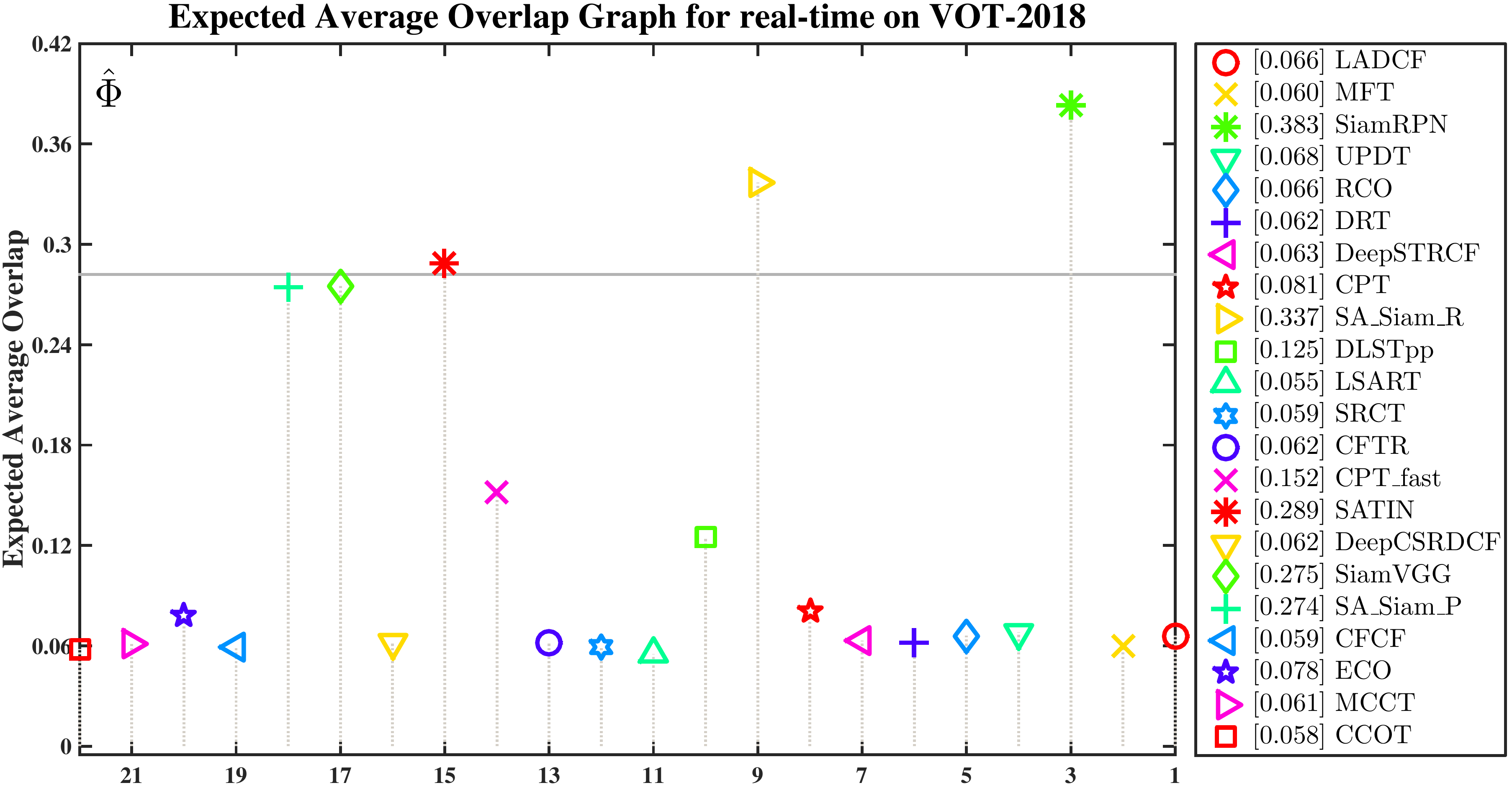}
\end{center}
\caption{Expected average overlap graph for the real-time experiment on the VOT-2018 challenge dataset~\cite{vot2018}.}
\label{fig:vot2018rt}
\end{figure}

For comprehensive evaluations, we further validate the tracking performance and efficiency of SATIN on the available VOT-2018 challenge dataset~\cite{vot2018}, which contains the same $60$ video sequences as the VOT-2017 challenge dataset. Two different experiments are conducted: a baseline experiment and a real-time experiment.  The baseline experiment is as same as the standard experiments in the VOT-2016 and VOT-2017 challenge datasets. While in the real-time experiment, if a tracker cannot estimate the tracking results in 40 ms, the previous estimated target object position and size will be considered as the current results straightforwardly. Fig.~\ref{fig:vot2018base} and Fig.~\ref{fig:vot2018rt} report the results of ours against 21 state-of-the-arts with respect to the EAO score in baseline and real-time experiments, respectively. Trackers whose EAO scores exceed 0.282 will be considered as state-of-the-arts~\cite{vot2018}. Although SATIN ranks the 15th with an EAO score of 0.294 in the baseline experiment, it achieves a third-best performance with the EAO score of 0.289 in the real-time experiment. Compared with the LADCF tracker~\cite{ladcf}, which delivers the best EAO score of 0.389 in the baseline experiment, SATIN outperforms it by $22.3\%$ in the real-time experiment. SiamRPN~\cite{siamrpn} obtains the best EAO score of 0.383 in the real-time experiment~\footnote{This is a particular optimized version of SiamRPN for VOT evaluations. The original version (which is exploited in the OTB evaluations as mentioned earlier) achieves an EAO score of 0.244 in both baseline and real-time experiments.}. However, it has more redundant hyperparameters, which cost much memory storage overhead~\cite{fasterrcnn,cornernet} when compared to SATIN. Overall, all the baseline and real-time results demonstrate the proposed approach has comparable performance and efficiency.

\subsection{Ablation studies}

\begin{table}[t]
\centering
\caption{Ablation studies of several variations of our tracker on OTB benchmarks using AUC and Precision metrics. The first, second and third best values are highlighted in \textbf{\textcolor[rgb]{1.00,0.00,0.00}{red}}, \textbf{\textcolor[rgb]{0.00,0.00,1.00}{blue}} and \textbf{\textcolor[rgb]{0.00,1.00,0.00}{green}}, respectively.}
\label{table:abla}
\begin{tabular}{p{2.2cm}p{2cm}<{\centering}p{2cm}<{\centering}p{0.1cm}<{\centering}p{2cm}<{\centering}p{2cm}<{\centering}}
\toprule
\multirow{2}{*}{Trackers} & \multicolumn{2}{c}{OTB-2015} & & \multicolumn{2}{c}{OTB-2013} \\ \cline{2-3}\cline{5-6}
                & \multicolumn{1}{c}{AUC~(\%)} & \multicolumn{1}{c}{Precision~(\%)} & & \multicolumn{1}{c}{AUC~(\%)} & \multicolumn{1}{c}{Precision~(\%)} \\
\midrule
SiamFC~\cite{siamfc}   & 57.8  & 76.7  & & 60.9  & 80.9      \\
\midrule
SATIN$_{NAA}$     & 60.8  & 81.2  & & 62.4  & 84.3      \\
SATIN$_{SENet}$   & 61.1  & 82.4  & & 63.2  & \textbf{\textcolor[rgb]{0.00,1.00,0.00}{87.1}}      \\
SATIN$_{NSA}$     & \textbf{\textcolor[rgb]{0.00,0.00,1.00}{62.6}}  & \textbf{\textcolor[rgb]{0.00,0.00,1.00}{83.4}}  &
                  & \textbf{\textcolor[rgb]{0.00,0.00,1.00}{64.5}}  & \textbf{\textcolor[rgb]{0.00,0.00,1.00}{87.6}}      \\
SATIN$_{NCA}$     & 61.5  & 82.2  & & 63.7  &86.5      \\
SATIN$_{NCP}$     & 57.2  & 79.1  & & 59.7  & 82.5      \\
SATIN$_{VGG}$     & \textbf{\textcolor[rgb]{0.00,1.00,0.00}{61.9}}  & \textbf{\textcolor[rgb]{0.00,1.00,0.00}{82.7}}  & & \textbf{\textcolor[rgb]{0.00,1.00,0.00}{64.3}}  & 86.4      \\
SATIN$_{MobN}$    & 59.3  & 80.5  & & 61.6  & 82.2      \\
\midrule
\textbf{SATIN}    & \textbf{\textcolor[rgb]{1.00,0.00,0.00}{64.1}}  & \textbf{\textcolor[rgb]{1.00,0.00,0.00}{84.4}}  & & \textbf{\textcolor[rgb]{1.00,0.00,0.00}{66.9}}  & \textbf{\textcolor[rgb]{1.00,0.00,0.00}{89.3}}      \\
\bottomrule
\end{tabular}
\end{table}

To validate our tracker and analyze the effectiveness of each component, several ablative variants of SATIN, as well as the baseline tracker SiamFC~\cite{siamfc}, are evaluated on OTB benchmarks. The detailed evaluation results are illustrated in Table~\ref{table:abla}.

Our full tracking approach (SATIN) outperforms all other variants. The effectiveness of the cross-attentional module is evaluated by comparison with three alternative variants, i.e., SATIN$_{NAA}$ (SATIN without all attention), SATIN$_{NSA}$ (SATIN without the spatial attention), and SATIN$_{NCA}$ (SATIN without the channel-wise attention). In addition, we implement a variant SATIN$_{SENet}$ that incorporates the recent SENet~\cite{senet} to learn channel-wise attention. SATIN$_{NSA}$ obtains absolute AUC and DP gains of $1.5\%$ and $1.0\%$ compared with SATIN$_{SENet}$. This is mainly attributed to the different pooling strategies of these two variants. Compared with SATIN$_{NAA}$, after integrating both channel-wise and spatial attentional information in the full approach, both robustness and accuracy are dramatically improved, with gains of almost $3.3\%$ and $3.2\%$ in AUC and DP scores, respectively. Without corner pooling operation (SATIN$_{NCP}$), inaccurate corner point detections significantly deteriorate the tracking results. This fact demonstrates that the corner pooling operations is more critical for detecting corner points. Specifically, we observe that the AUC score of SATIN$_{NCP}$ is inferior to that of the baseline SiamFC with a $1.2\%$ drop. But the DP score is around $2.4\%$ higher than that of SiamFC owing to that centroid point detection does not need corner pooling operation.

To investigate how the proposed lightweight hourglass network contributes to SATIN, we first implement the variant SATIN$_{VGG}$, which stacks two VGG-M~\cite{vgg} as the backbone network. Compared with the full SATIN tracker, the performance of SATIN$_{VGG}$ drops more than $2.6\%$ and $2.9\%$ in terms of AUC and DP scores on the OTB-2015 benchmark dataset. We also construct another variant named SATIN$_{MobN}$ by replacing the lightweight hourglass networks with two MobileNets~\cite{mobilenet}. However, this replacement causes a significant performance drop of $5.3\%$ from $66.9\%$ to $61.6\%$ in the AUC score on the OTB-2015 benchmark dataset. These results~\footnote{For comprehensive and fair evaluations, we also add some convolutional and upsampling layers following each original VGG-M and MobileNet to make the output feature maps at the desired resolution in SATIN$_{VGG}$ and SATIN$_{MobN}$, respectively.} prove that our backbone network can learn more contextual representations than the off-the-shelf and lightweight networks. For comprehensive studies on the effectiveness of the proposed lightweight hourglass network, we further compare it with the original hourglass network~\cite{hourglass} and the  MobileNet~\cite{mobilenet}. Table~\ref{table:para} summarizes the experimental results. We observe that using deeper backbone networks for feature extraction may lead to further performance improvement but also with the extra computational burden. Although the original hourglass network slightly outperforms the proposed lightweight hourglass network on both OTB-2015 and VOT-2017 datasets, it is much larger and complicated than the latter. Specifically, compared with the single-stage original hourglass network ($\times$1), the \#Params and \#GFLOPs of the single-stage lightweight hourglass network ($\times1$) are much lower and less than $35\%$ while the speed is $15\times$ fast. It is worth noting that adding another lightweight hourglass network in the backbone, i.e., the two-stage lightweight hourglass network ($\times2$), significantly improves performance. The AUC score on the OTB-2015 benchmark dataset is enhanced by $3.0\%$ from 0.611 to 0.641, and the EAO score on the VOT-2017 challenge dataset is improved by $3.2\%$ from 0.262 to 0.294, respectively, at the costs of $174\%$ \#Params and $161\%$ \#GFLOPs. Moreover, the two-stage lightweight hourglass network ($\times2$) performs better than the single-stage original hourglass network ($\times1$) while significantly being $2.9\times$ smaller and $2.4\times$ less computational cost, which evidences reasonable complexity and performance tradeoff is critical for visual tracking.

These results indicate that each component brings individual improvement, and all of them work together to achieve surprisingly excellent tracking performance.

\begin{table}[t]
\centering\footnotesize
\caption{Effect on the depth of backbone networks in SATIN.}
\label{table:para}
\begin{tabular}{rcccccc}
\toprule
\# \begin{tabular}[c]{@{}r@{}}Backbone Networks\end{tabular} &
\begin{tabular}[c]{@{}c@{}}MobileNet\\($\times$1)\end{tabular} &
\begin{tabular}[c]{@{}c@{}}MobileNet\\($\times$2)\end{tabular} &
\begin{tabular}[c]{@{}c@{}}Original\\Hourglass\\($\times$1)\end{tabular} &
\begin{tabular}[c]{@{}c@{}}Original\\ Hourglass\\ ($\times$2)\end{tabular} &
\begin{tabular}[c]{@{}c@{}}Lightweight \\Hourglass\\ ($\times$1)\end{tabular} &
\begin{tabular}[c]{@{}c@{}}Lightweight \\Hourglass\\ ($\times$2)\end{tabular} \\
\midrule
AUC on OTB-2015 &0.577&0.593&0.632&0.650&0.611&0.641\\
EAO on VOT-2017 &0.252&0.270&0.287&0.302&0.262&0.294\\
Speed (fps)           &93&77&9&$\sim$2&41&27\\
\# Params (M)   &5.3&9.8&32.2&57.8&11.3&19.7\\
\# GFLOPs        &0.87&1.43&36.2&66.5&15.4&24.8\\
\bottomrule
\end{tabular}
\end{table}

\section{Conclusions}\label{sec:5}

In this paper, we propose a new deep architecture named SATIN for high-performance visual tracking. It utilizes a Siamese lightweight hourglass network as the backbone network, which can extract more global and local contextual representations at multiple scales in a repeated bottom-up and top-down manner. We model generic visual tracking as a set of keypoint detection tasks, i.e., SATIN tracks any target object by detecting three keypoints of its bounding box, including the top-left corner point, the centroid point, and the bottom-right corner point. Thus, our approach eliminates the need for designing anchor boxes or multi-scale pyramids. Meanwhile, a cross-attentional module is employed to heuristically learn where to emphasize or suppress along channel-wise and spatial dimensions on object representations, therefore the tracking performance is promoted further. Without bells and whistles, our approach achieves state-of-the-art results on four popular benchmark datasets. Furthermore, SATIN runs at $27$ fps, which is far above real-time frame rate.

\section*{Acknowledgments} \label{sec:ack}

This work was supported by the National Natural Science Foundation of China under Grant No.~31701187, the Science and Technology Planning Program of Guangdong Province under Grant No.~2016B090918047, and Promotional Credit from Amazon Web Service, Inc. The authors would like to thank all the anonymous reviewers for their insightful comments and suggestions that have helped to significantly improve the quality of this paper.

\bibliography{arxiv}

\end{document}